\documentclass[12pt]{article}

\usepackage[margin=1in]{geometry}
\usepackage{setspace}
\usepackage{hyperref}
\usepackage{microtype}
\usepackage{csquotes}
\usepackage{enumitem}
\usepackage{amsmath,amssymb}
\usepackage{subcaption}
\usepackage{booktabs} 
\usepackage{multirow}
\usepackage{microtype}
\usepackage{graphicx}
\usepackage{array}

\graphicspath{{img}}

\title{Do Transformers Understand Ancient Roman Coin Motifs Better than CNNs?}
\author{David Reid and Ognjen Arandjelovi\'c\\
School of Computer Science\\
University of St Andrews\\
Scotland KY16 9AJ, UK
}
\date{}

\begin{document}
\maketitle

\begin{abstract}
Automated analysis of ancient coins has the potential to help researchers extract more historical insights from large collections of coins and to help collectors understand what they are buying or selling. Recent research in this area has shown promise in focusing on identification of semantic elements as they are commonly depicted on ancient coins, by using convolutional neural networks (CNNs). This paper is the first to apply the recently proposed Vision Transformer (ViT) deep learning architecture to the task of identification of semantic elements on coins, using fully automatic learning from multi-modal data (images and unstructured text). This article summarises previous research in the area, discusses the training and implementation of ViT and CNN models for ancient coins analysis and provides an evaluation of their performance. The ViT models were found to outperform the newly trained CNN models in accuracy.
\end{abstract}

\section{Introduction}

\emph{Numismatics} is the study of physical currency, such as coins. Ancient numismatics, the focus of this paper, concerns ancient coins (such as Ancient Greek, Roman, and Celtic) and is of interest to both collectors and scholars. Over the course of the last 15 years or so, work on the employment of computer vision (CV) and machine learning (ML) in ancient coin analysis has been attracting an increasing attention from the research community, owing both to its potential to save time and increase the availability of emergent information to scholars, and assist as a learning tool to hobbyists. The task has also proven to be a most difficult one, introducing a series of domain-specific CV and ML challenges unencountered in the more mainstream image understanding applications.

While most of the existing research in this realm thus far has focused on image-to-image matching problem, in the present paper we build upon a new strand of methods that seek to describe the semantic content depicted on coins, that is, that aim to understand their stylized visual elements. This shift in focus is driven by the observation that the volume of possible ancient coin issue types is enormous and practically prohibitive to obtain for the aforesaid matching, motivating more nuanced analysis more akin to how a human expect would approach it. In particular, following on the Convolutional Neural Network (CNN) approach introduced by Cooper and Arandjelovi\'c~\cite{cooper_learning_2020}, herein we investigate if the use of the Vision Transformer (ViT), an architecture which has largely overtaken the popularity of CNNs for many visual tasks, confers benefit in the recognition of such nuanced content.

The remainder of the article is organized as follows. Firstly, considering the interdisciplinary nature of our work, we briefly summarize the key technical terminology so as to facilitate the understanding of the article content by as wide of an audience as possible. In Section~\ref{s:prior_work} we survey the relevant prior work, starting from work on the use of ML and CV in ancient numismatics, continuing to a survey of the technical underpinnings of the Transformer architecture and then the Vision Transformer in particular, finishing with a discussion of research on explainable AI. Then we proceed with a comprehensive description of the proposed methodology in Section~\ref{s:proposed}. In Section~\ref{s:results} we present a description of our empirical evaluation, including the details of the data sets used, the obtained results, and a discussion thereof. Finally, in Section~\ref{s:conc} we make concluding remarks, highlight the potential limitations of the present research, and make recommendations for future work.

\subsection{Relevant specialist terminology}
In this context, a \emph{coin} refers to an individual physical specimen, whereas its \emph{issue} is the type of the coin and is of primary interest in many classification tasks. For two coins to be considered of the same issue, their designs must match in terms of the semantic content on either side, though there can be minor variations in artistic depictions and the spacing between letters, which complicates automated analysis. The two sides of a coin, colloquially known as `heads' and `tails', are technically called the \emph{obverse} and \emph{reverse} respectively. Inscriptions around the edge of a coin are referred to as a \emph{legend}. The obverse (heads) features a bust as the central motif and the surrounding legend provides the name and titles of the issuer. The reverse has a central motif that may be a person, place or object, which is often related to the reverse legend.

\section{Related Work}\label{s:prior_work}

The primary focus of this research is the application of the Vision Transformer (ViT) architecture to the task of automated analysis of ancient coins. To provide the necessary context for the subject matter, this section consists of: a survey of recent work to automate ancient coin analysis; an overview of Vision Transformers and related work; and a brief overview of visual explanation of deep learning models.

\subsection{Automated Ancient Coin Analysis}

Historically, analysis of ancient coins has been performed by experts with long-acquired specialist knowledge, often with reference to authoritative sources for common classification schemes \cite{crawford_roman_1974,mattingly_roman_1923}. This is time-consuming and costly; therefore, efforts to automate such analysis have increased dramatically as computer vision has become ubiquitous in modern society.

Arandjelovi\'c and Zachariou \cite{arandjelovic_images_2020} identified eight key problems in the application of computer vision and machine learning to ancient numismatics, which are far from completely solved. These included specimen matching (for tracking an individual coin), issue matching (for identifying a coin's type) and forgery recognition. Forgery recognition and specimen matching are of obvious interest to collectors and dealers of coins, who want to be sure of exactly what they are trading.

\subsubsection{Challenges}

A number of factors make the automated analysis of ancient coins in images difficult and therefore an interesting research problem. Firstly, there is a high level of intra-class variability for most analysis problems. This can be due to differences in the dies used to produce coins of the same issue, which may differ in artistic depictions of motifs or spacing in legends, or variability introduced by the degradation of coins, both from wear while they were in circulation and, later, from the environment in which they were stored or buried prior to excavation. Secondly, there are many types of ancient coins (at least 43,000 Roman imperial coin issues~\cite{_online_coins} and 1,900 republican ones~\cite{anwar_deep_2021}). Many of these coin issues do not have readily available high-quality images, which makes visual coin matching in general impractical, as noted by Cooper and Arandjelovi\'c \cite{cooper_learning_2020}. Further challenges are posed by the lack of meaningful textural information from coins \cite{arandjelovic_images_2020} and the variability among photos of coins, which can have uneven lighting or include objects irrelevant to analysis tasks.

\subsubsection{Prior Work on Automation}

Automated analysis of ancient coins is a research field that has grown considerably in the past decade. During that time, a range of methods have been applied.

Earlier work \cite{zaharieva_image_2007,kampel_recognizing_2008,arandjelovic_reading_2012} tended to focus on visual matching of coin issues based on local features, typically using scale-invariant feature transform (SIFT) \cite{lowe_distinctive_2004}. Such techniques, which worked well for modern coins, were found to perform poorly on ancient coins, due to excessive wear and irregularities in the designs \cite{zaharieva_image_2007}. The main issue with just using local features is that the spatial relationships between features, which are important for analysis, are lost. Some success was found by using locally-biased directional kernel features (which uses SIFT) to capture some of this spatial information \cite{arandjelovic_automatic_2010}, but the overall accuracy of classification (57\%) was still far too low to be of practical use. Spatial tiling schemes are often used along with local features to loosely capture some of the spatial information between features. Anwar et al.\ \cite{anwar_ancient_2015} developed a model, based off SIFT features, that applied a circular tiling scheme and captured triangular geometric information between identical visual `words'. This was shown to be superior to a simple bag-of-visual-words model (which did not capture any spatial relationships) and invariant to image rotations, scale changes and translations. The overall accuracy achieved (80\%) was still not fit for practical purposes, but the method demonstrated the importance of including spatial relationships between features in the analysis of ancient coins.

With the explosion in the use of deep learning, more recent work has explored the application of deep Convolutional Neural Networks to ancient numismatics \cite{schlag_ancient_2017, cooper_learning_2020, anwar_deep_2021, he_deep_2016, zachariou_visual_2020}. This work is of particular interest, as prior to the development of the Vision Transformer architecture, CNNs had become the dominant tool of choice for most computer vision tasks. Cooper and Arandjelovi\'c~\cite{cooper_learning_2020} argued that the large number of ancient coin issues relative to the number of readily available images of them rendered most visual matching tasks impractical.\ Instead, they argued that research efforts should be focused on understanding individual semantic elements depicted in coins, as these are often shared among multiple coin issues and so can feasibly be learnt, but they are still specific enough that their identification would narrow the set of possible matches for a coin. Their work used a data set of 100,000 descriptions and images of coins from online auction lots, and is the same data set used for this work, so it is of particular interest as a benchmark. It combined the use of unstructured text analysis of coin descriptions with visual learning using a CNN to identify five common semantic concepts depicted on ancient coins, achieving accuracy rates of around 80\%.

Whereas much of the aforementioned work focused on the reverse of coins, which tend to be more visually distinct than the obverse sides, and relied on little domain-specific information, some work has focused on the obverse side and made use of domain-specific information. Arandjelovi\'c et al.\ extracted the obverse legend and used this to search for images of potential matches, which were then visually compared~\cite{arandjelovic_reading_2012}. For future work, they proposed using domain-specific features over SIFT features. Later work by Schlag and Arandjelovi\'c~\cite{schlag_ancient_2017} noted that the fine detail of legends is often lost by wear, rendering the previous approach of Arandjelovi\'c impractical~\cite{arandjelovic_reading_2012}. Their own work used deep Convolutional Neural Networks to recognise emperors' profiles on the obverse side and demonstrated top-1 recognition rates of around 85\% and top-3 recognition rates of around 95\%. Anwar et al.~\cite{anwar_deep_2021} proposed a model (CoinNet) that used Compact Bilinear Pooling with a CNN to classify coins from a new, public data set (RRCD) of 18,000 images featuring 228 reverse motif classes. To demonstrate their model's generalization ability, they tested it on a disjoint data set of unseen coin types, achieving accuracy scores of 68-96\% in identifying the central reverse motif. Although CNNs are invariant to image translations, they typically perform poorly for image rotations. He and Wu \cite{he_ancient_2021} proposed the Rotation Transformation Network (RTN) to solve this problem. During training, the Rotation Transformation Module (RTM - a CNN) is presented with rotated images of coins and learns to predict angles of rotation. After training, the RTM is able to rotate coin images back to their canonical orientation before they are passed to a classification module, thus giving the overall network rotational invariance.

In a recent paper, Aslan et al.\ \cite{aslan_two_2020} applied a game theoretic, graph-based algorithm (Graph Transduction Games) for semi-supervised learning of coin classifications, demonstrating improved accuracy when images of both sides of the same coin were used in classification. They also noted that, in the literature, each research group created their own coin data sets due to a lack of a publicly available large-scale data set for ancient coin classification, which makes comparison of works more difficult. As noted by Aslan et al., a few data sets have been made publicly available upon request, but it seems that there is still little crossover of data sets among research groups. We would encourage future research to report findings on existing public data sets where possible, so that findings can be readily compared and replicated.

While the aforementioned research has focused on improving classification accuracy, some other investigations have been performed. Fare and Arandjelovi\'c \cite{fare_ancient_2017} examined the impact of coin grading on classification accuracy, which has received little attention in the literature despite the obvious impact that a coin's condition can have on its identifiability. Although grading had some measurable impact on accuracy, overall they found the SIFT-based methods that they had employed yielded poor results. Zachariou and Arandjelovi\'c \cite{zachariou_visual_2020} developed a generative adversarial network (GAN) to virtually restore a coin's visual detail that has been lost through degradation, which produced some convincing results, but the lack of ground truth against which to compare results meant evaluation was very subjective.

There is still much that has not yet been explored in the literature. For example, to date there has not been any work on forgery recognition or die-matching, as noted in Arandjelovi\'c and Zachariou~\cite{arandjelovic_images_2020}. While there has been a lot of focus on the visual matching of coin issues, there has been less work focused on extracting semantic content from previously unseen coin issues. Herein we advance efforts in this realm by adopting the Vision Transformer architecture motivated by its successes in a wide range of visual tasks yet thus far only once employed in the context of ancient coin analysis~\cite{guo2023siamese}.

\subsection{Transformer Architectures}

This section provides an overview of the Vision Transformer (ViT) architecture and the Transformer architecture, which it is based on.

\subsubsection{Transformer}\label{s:transformer}

The Transformer is a deep learning architecture for natural language processing (NLP) applications, which, after its publication in 2017 by a team from Google Brain~\cite{vaswani_attention_2017}, revolutionised the field of NLP and led to new state-of-the-art models for many NLP tasks, while also reducing training times for large data sets. Google's `Bidirectional Encoder Representations from Transformers' (BERT) model~\cite{devlin_bert_2019a} has been used to improve Google's search functionality for more complex queries. OpenAI's `Generative Pre-trained Transformer 3' (GPT-3) model~\cite{brown_language_2020a} became the largest neural network ever made and made headlines with its impressive ability to generate text that appeared to have been written by humans.

The Transformer follows a similar encoder-decoder architecture to previous models, in which one sequence of tokens, representing words in a sentence, is used to generate another sequence (e.g. a translation of the sentence). What is special about the Transformer architecture is that, unlike it predecessors, it does not use convolutional layers or recurrent connections but instead largely relies on \emph{self-attention}, as suggested by the paper's title: `Attention is All You Need' \cite{vaswani_attention_2017}. In this context, \emph{attention} refers to a general mechanism for focusing on information relevant to the current task. In a neural network, an attention unit's role is to map equal length sequences of \emph{query}, \emph{key} and \emph{value} vectors to a sequence of \emph{context} vectors, each of which is a weighted mean of the value vectors, weighted towards those that are most relevant to the corresponding position in the sequence for the given task. The model used for computing these attention weights uses three matrices, which are learnt during training:\\
\begin{align}
    &\mathbf{Q} \in \mathbb{R}^{n \times d_k}\textnormal{, whose rows are query vectors}\\
    &\mathbf{K} \in \mathbb{R}^{n \times d_k}\textnormal{, whose rows are key vectors}\\
    &\mathbf{V} \in \mathbb{R}^{n \times d_v}\textnormal{, whose rows are value vectors}
\end{align}
Here $n$ is the maximum sequence length, $d_k$ is the dimensionality of the query and key vectors and $d_v$ is the dimensionality of the value vectors.
In a translation context, the query vectors correspond to words in the target language, whereas the key and value vectors would correspond to words in the source language. Let the words of the input sentence be represented by the rows of $\mathbf{X} \in \mathbb{R}^{n \times d}$, then the learnable embeddings $\mathbf{W}^K \in \mathbb{R}^{d \times d_k}$ and $\mathbf{W}^V \in \mathbb{R}^{d \times d_v}$ project the input $\mathbf{X}$ to the key matrix $\mathbf{K}=\mathbf{X}\mathbf{W}^K$ and the value matrix $\mathbf{V}=\mathbf{X}\mathbf{W}^V$. Let the words of the translated output up to the current token be represented by $\mathbf{Y} \in \mathbb{R}^{n \times d}$, then a learnable embedding $\mathbf{W}^Q \in \mathbb{R}^{d \times d_k}$ projects $\mathbf{Y}$ to the query matrix $\mathbf{Q}=\mathbf{Y}\mathbf{W}^Q$.

In the Transformer architecture, `Scaled Dot-Product Attention' is used: how much the $j$th value vector contributes to the $i$th context vector is determined by the dot-product of the $i$th query vector and $j$th key vector. The dot-products are scaled by $\frac{1}{\sqrt{d_k}}$, lest they become too large, resulting in problematically small gradients. Softmax is applied to the dot products to obtain positive weights that sum to one. The attention weights are then multiplied by the value vectors to obtain the context vectors. In matrix form, the function of a single attention unit can be written as:
$$\textnormal{Attention}(\mathbf{Q}, \mathbf{K}, \mathbf{V}) = \textnormal{softmax} \left(\frac{\mathbf{Q} \mathbf{K}^T}{\sqrt{d_k}}\right)\mathbf{V}$$
A \emph{self}-attention unit is just an attention unit in which the input matrices are all derived from a single input sequence $\mathbf{X}$, such that it computes $\textnormal{Attention}(\mathbf{X}\mathbf{W}^Q,\mathbf{X}\mathbf{W}^K, \mathbf{X}\mathbf{W}^V)$, whereas a general attention unit computes $\textnormal{Attention}(\mathbf{Y}\mathbf{W}^Q,\mathbf{X}\mathbf{W}^K, \mathbf{X}\mathbf{W}^V)$, where the query matrix is computed from a different sequence ($\mathbf{Y}$) to the one used ($\mathbf{X}$) to compute the key and value matrices. Intuitively, a self-attention unit returns context vectors that represent the internal structure of the input sequence and the dependencies between different parts.

Previous language models that used Recurrent Neural Networks (RNNs) struggled to learn long-range dependencies between words in long sequences, because they processed tokens sequentially, meaning that any state passed forward in the network had to encode the entire sequence up to the current token, which became less effective the longer the sequence was. In the Transformer model, the self-attention mechanism operates over the entire sequence of input symbols, so it is equally able to handle dependencies over any range. Another issue with sequential processing is that training could not be parallelized as effectively. The Transformer employs \emph{Multi-Head Self-Attention} (MSA), running multiple identical, but separately parametrized, self-attention units (`heads') in parallel. This allows it to attend to different regions for different representations concurrently, which would not be possible with a single head, as the weighted mean of many many points of interest would result in a lack of focus on anything in particular. The output of an MSA unit is the concatenation of the output vectors from the individual heads, projected by a matrix back to vectors of dimensionality $d_\textnormal{model}$, which is constant throughout the Transformer.

The encoder in the Transformer consists of $N=6$ identical layers. The input to the decoder is an embedding of the input sequence (e.g. the sentence to be translated) and the output feeds into the decoder. Each layer in the encoder has two sub-layers: the first is an MSA unit; the second is a fully connected feed-forward network (FFN), consisting of two linear transformations with a ReLU activation ($ReLU(x) = \max(0,x)$) in between. Skip connections around each sub-layer are used. These are a widely used feature in deep learning that help with the `vanishing gradient problem' that would otherwise occur in deeper networks, whereby the calculation of the product of many small gradients results in very slow training times. Furthermore, skip connections have been show to improve a network's ability to learn by flattening the loss landscape \cite{li_visualizing_2018}. As well as skip connections, layer normalization is applied to each sub-layer, as this has been show to improve training times \cite{ba_layer_2016}. 

The Transformer decoder consists of $N=6$ identical layers, containing three sublayers. As for the encoder, an MSA unit and a FFN are two of the sub-layers, and skip connections and layer normalization are used. The extra sub-layer is a Multi-Head Attention unit (not a Multi-Head \emph{Self}-Attention unit), for which the key and value matrices come from the output of the encoder and the query matrix comes from the output of the preceding MSA unit. This is how information flows from the encoder into the decoder. The input to the decoder is the tokens of the output sequence (e.g. the translated sentence).

Unlike the structure of an RNN or a CNN, the Transformer architecture does not implicitly contain any notion of position for the input data. Instead, positional encodings for each token are added to the input embeddings that are passed into the encoder. Whereas an RNN or a CNN has a strong inductive bias towards locality, a Transformer has few inductive biases, so it must learn the significance of positional relationships during training. This lack of a strong inductive bias makes Transformers very generic and able to model long-range dependencies, but at a cost of worse performance for small training sets, for which sensible inductive biases can lead to better performance. Since Transformer needs a large amount of training data to outperform other models, transfer learning is typically used: a Transformer model that has been pre-trained on a very large data set is fine-tuned with training data for a specific task, enabling it to make use of previously learnt generalisations and thus avoiding the need to train from scratch for each task.

It is worth noting that despite the title of `Attention is All You Need'~\cite{vaswani_attention_2017}, a later paper titled `Attention is Not All You Need: Pure Attention Loses Rank Doubly Exponentially with Depth' showed that self-attention has a strong inductive bias towards token uniformity, and the skip connections and feed-forward networks in the Transformer architecture are crucial to its ability to learn complex models \cite{dong_attention_2021a}.

\subsubsection{Vision Transformer}

\begin{figure}[h]
    \centering
    \includegraphics[width=0.7\textwidth]{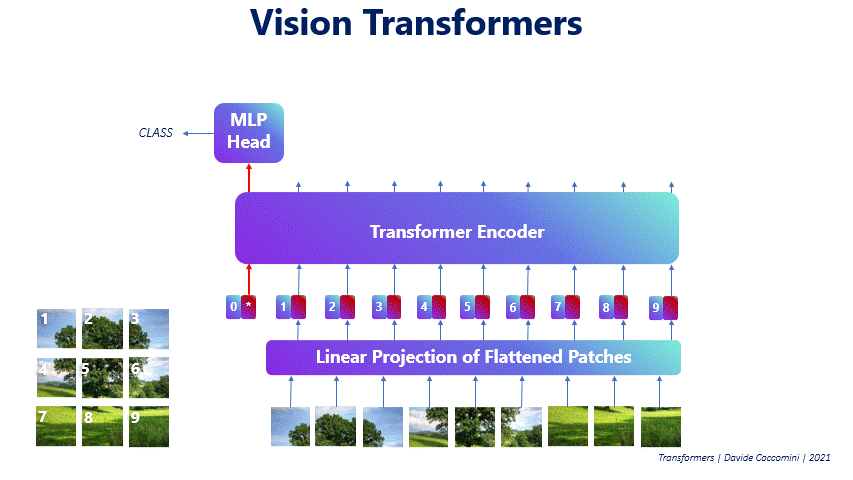}
    \caption[ViT architecture]{A high-level view of the ViT architecture, showing how it is built on top of a standard Transformer Encoder. Image source: Davide Coccomini. Licensed to share under CC-BY-SA-4.0.}
    \label{fig:vit-architecture}
\end{figure}

The Vision Transformer (ViT) architecture \cite{dosovitskiy_image_2020} is a direct descendant of the Transformer architecture described in the previous section (\ref{s:transformer}). Its creators followed the original Transformer architecture as closely as possible, enabling existing efficient Transformer implementations to be used with relative ease. Whereas Transformer was designed for sequence-to-sequence language tasks and therefore had an encoder and a decoder, ViT is used for image classification tasks and so it only has an encoder, to which tokens representing an image are provided as input. The ViT encoder is a standard Transformer encoder (Figure \ref{fig:vit-architecture}).

One of the main design decisions for ViT was how to embed the image. A naive implementation of self-attention would allow each pixel to attend to every other pixel, resulting in $O(n^2)$ asymptotic time and space complexity for images of $n$ pixels. This would be prohibitively expensive, so a simplification has to be made. In the case of ViT, the simplification is to use image patch embeddings as the input tokens rather than pixels. Each image of width $W$ and height $H$ is divided into patches of $P\times P$ pixels, resulting in $N=WH/P^2$ patches, which is a small enough number to make self-attention across patches feasible. The pixel data of the patches are flattened to vectors, which are projected by a learnable embedding to vectors of dimensionality $d_model$, which is the size of the vectors used throughout the layers of the encoder.

As ViT is used for classification tasks, an additional learnable class token embedding is passed in to the encoder as the zeroth `patch embedding'. After the final layer of the encoder, an additional FFN is added, which maps the context vectors from the zeroth position in the last layer to the image classes. During training, the network learns to encode in these output context vectors a representation of the image that is then used for classification purposes. During fine-tuning, a different FFN is used to project this image representation to the classes specific to the problem domain. In the FFNs of the encoder, GELU is used as the activation function, whereas Transformer uses ReLU, but the authors offered no explanation for this modification.

As for Transformer, ViT includes positional information in the data passed to the encoder, otherwise spatial relationships between patches could not be learnt. The authors found that a 2D-aware positional embedding offered no significant improvement over a 1D positional embedding, so they used a 1D positional embedding and left it to the network to learn how the patch positions were spatially related to one another.

In their evaluation of ViT, Dosovitskiy et al.~\cite{dosovitskiy_image_2020} compared ViT to CNNs, specifically ResNets~\cite{he_deep_2016}, and hybrids of ViT and CNNs, for which the input sequence to the ViT was formed from the feature maps of a trained CNN, rather than image patches. While the hybrids outperformed ViT for smaller data sets, (presumably, because the features already encoded at least local structure within the data), this performance difference vanished for larger data sets, demonstrating the ability of ViT to learn complex features without a strong inductive bias towards local features. As the size of the training data set was scaled up to 300 million images, the performance of ViT continued to increase without reaching a saturation point, showing that more data is better when it comes to training ViT models.

For CNNs, the size of dependencies that can be represented by a feature at a given layer is limited by the receptive field for the feature. The size of the receptive field increases with depth. In contrast, ViTs can model long-range dependencies in their lowest layers. By visualising the mean distance in image space over which information was integrated for a given layer of ViT, Dosovitskiy et al.\ found that some heads even in the lowest layers of ViT modelled long-range dependencies, whereas others were highly localised. For the hybrid models tested, highly localised attention was less pronounced, suggesting that the role played by the highly localised attention heads was similar to that played by early convolutional layers in a CNN.

\subsection{Visual Explanations}
\label{sec:visual-explanations}

One of the current challenges for deep learning researchers and practitioners is that deep learning models can be incredibly complex and so their output predictions can be difficult to understand. This creates a problem of trust, especially in safety critical environments (e.g. medical diagnosis \cite{valsson_nuances_2022}), in that users of such models may be unwilling to deploy them to tackle real-world problems, for lack of confidence that the model will not produce harmfully inaccurate predictions in some circumstances, due to a bias in the model. XAI (eXplainable Artificial Intelligence) research tries to address this challenge by providing tools and techniques for helping people to interpret model outputs. For computer vision problems, visual explanations are often the most intuitive. Input images overlaid with additional information from an XAI tool can offer insight into what a model understands of a particular image, enabling the tool users to see whether there are visual clues indicating that the model has a poor understanding of the image, potentially indicating a more general problem with the model.

In broad terms, XAI tools can be divided into `white-box' tools and `black-box' tools, depending on whether the tool requires access to the internals of the model in question. 

Grad-CAM \cite{selvaraju_gradcam_2017} is a popular white-box tool for CNNs, which uses backpropagation to calculate gradients of the last convolutional layers of a CNN with respect to the model's output classes. This enables it to produce a class-specific heatmap that shows how the activations in the final layers of the CNN relate to a specific class. For a given image and class, the `hot' parts of the heatmap should correspond to where an object of the class may be found.

Since black-box tools do not have access to the internals of the model under analysis, they are more restricted in what information they can convey about the model. Furthermore, since the only information they can obtain is from passing different inputs into the model and observing the impact that this has on the model's output, the tools can sometimes take longer to run, compared to white-box tools that can directly extract parameter values and intermediate result from models. What makes black-box models attractive is that they can be applied to any model without modification of the model. 

HiPe \cite{cooper_believe_2022} is a black-box tool that efficiently generates class-specific saliency maps, which highlight the regions of an image that support a positive prediction for a particular output class. HiPe uses a number of binary, rectangular masks of decreasing size, which correspond to particular regions of the input image. For each of the generated masks, a copy of the input image is made in which the masked region is perturbed (the default behaviour is that HiPe replaces the pixel values in the masked region with the mean pixel value for the masked area) and this modified input is passed to the model. If the model's predicted output for a particular class is lower for the perturbed image, then the region is salient to a prediction for that class. The magnitude of the change in the model's output determines just how salient the region is. One of the key strengths of HiPe is that it can efficiently generate saliency maps without user configuration of feature sizes. It starts by using larger perturbation masks and then successively maps to a finer level of detail image regions that are of greater saliency to the model and class in question.

\section{Proposed Approach}\label{s:proposed}

The focus of this paper was to compare the performance of two computer vision neural network architectures on the task of analysis of semantic elements on ancient coins. The main steps in this work were as follows:

\begin{enumerate}
    \item Preprocess the raw data, extracting the relevant part of each image and labelling it per semantic element of interest based off the image's description.
    \item Load and train models for two very different deep learning architectures (ViT and CNN) on each of the semantic elements.
    \item Evaluate the models per semantic element, comparing their performance both quantitatively, based on importance performance statistics, and qualitatively, assisted by saliency maps generated for images of interest.
\end{enumerate}

Additionally, a small software module was developed for the automatic identification of common semantic elements depicted by the coins in the data set, based on the image descriptions. The purpose of this was to identify concepts that models could subsequently be trained on.

\subsection{Data Preparation}
\label{sec:impl-data-prep}

As previously noted, the raw data set for this work consisted of images and descriptions of ancient coins from a coin auction aggregator website; the data set is described in detail in Section~\ref{ss:data}. It was necessary to preprocess this raw data set to generate a new data set, consisting of labelled images of only the reverse of each coin, which could then be used for training the models. The preprocessing of the images and the labelling of them based on their descriptions were carried out independently of each other.

\begin{figure}[t]
    \centering
    \includegraphics[width=10cm]{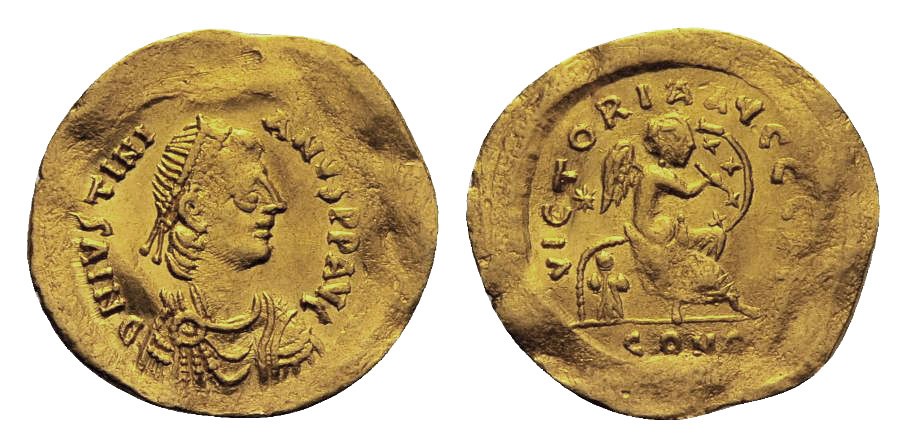}
    \begin{verbatim}
Justinian I AV Semissis. Constantinopolis, AD 527-565. D N IVSTINIANVS P P
AVG, diademed, draped and cuirassed bust right / VICTORIA AVGGG, Victory
seated right, inscribing numerals on shield; star to right, [christogram
to left]. MIBE 17; Sear 143. 2.15g, 19mm, 6h.  Near Extremely Fine.\end{verbatim}
    \caption[Example of data set sample]{An example of a typical sample from the data set, consisting of an image that shows a coin's obverse on the left and its reverse on the right, and a description of the coin.}
    \label{fig:coin-5}
\end{figure}

Most of the images in the raw data set showed both sides of a single coin side-by-side, with the obverse on the left and the reverse on the right; see Figure~\ref{fig:coin-5}. Since the reverse sides of ancient coins are usually richer in semantic motifs relevant to analysis, they were the focus of this paper, building upon the work of \cite{cooper_learning_2020}. As in the work of Cooper and Arandjelovi\'c~\cite{cooper_learning_2020}, the desired format for the data set was for each image to include only the reverse side of a single coin, thus reducing the amount of irrelevant data passed to the models, and to have binary labels for each of the semantic elements of interest, indicating whether they were present on the coin. Ideally, the image labels would also include spatial information as to the location, size and shape of the associated elements (e.g. labels per pixel), but obtaining such fine-grained labelling would require a prohibitive amount of human expert time and effort, so instead we make do with \emph{weak supervision}, in which labels apply at the granularity of a whole image.

\subsubsection{Data Preprocessing}

\begin{figure}[t]
    \centering
    \begin{subfigure}[b]{0.4\textwidth}
        \centering
        \includegraphics[width=\textwidth]{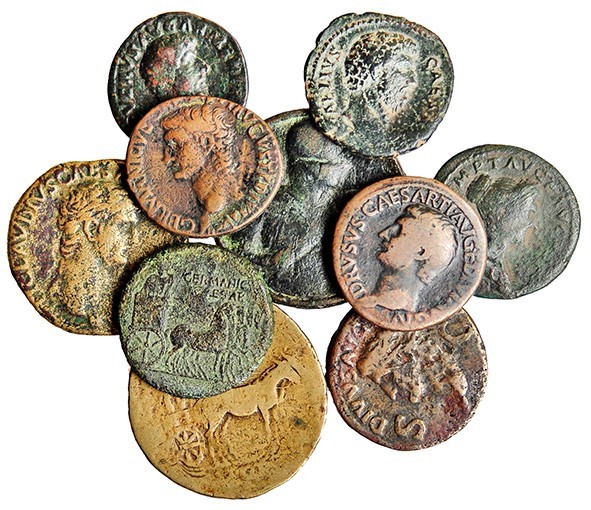}
        \caption{Many overlapping coins}
    \end{subfigure}
    \hfill
    \begin{subfigure}[b]{0.4\textwidth}
        \centering
        \includegraphics[width=\textwidth]{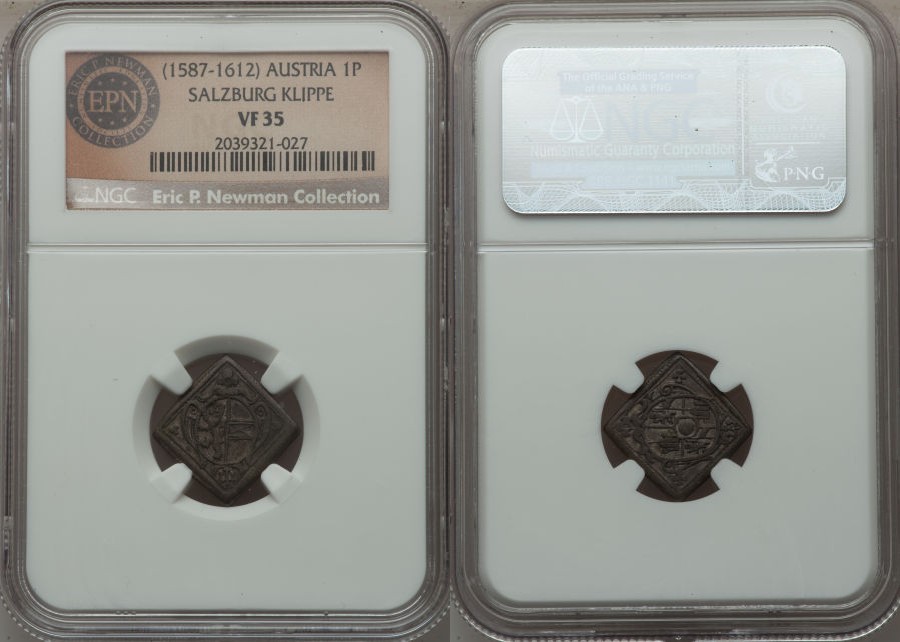}
        \caption{Coins in a case}
    \end{subfigure}

    \vspace{10pt} 
    
    \begin{subfigure}[b]{0.36\textwidth}
        \centering
        \includegraphics[width=\textwidth]{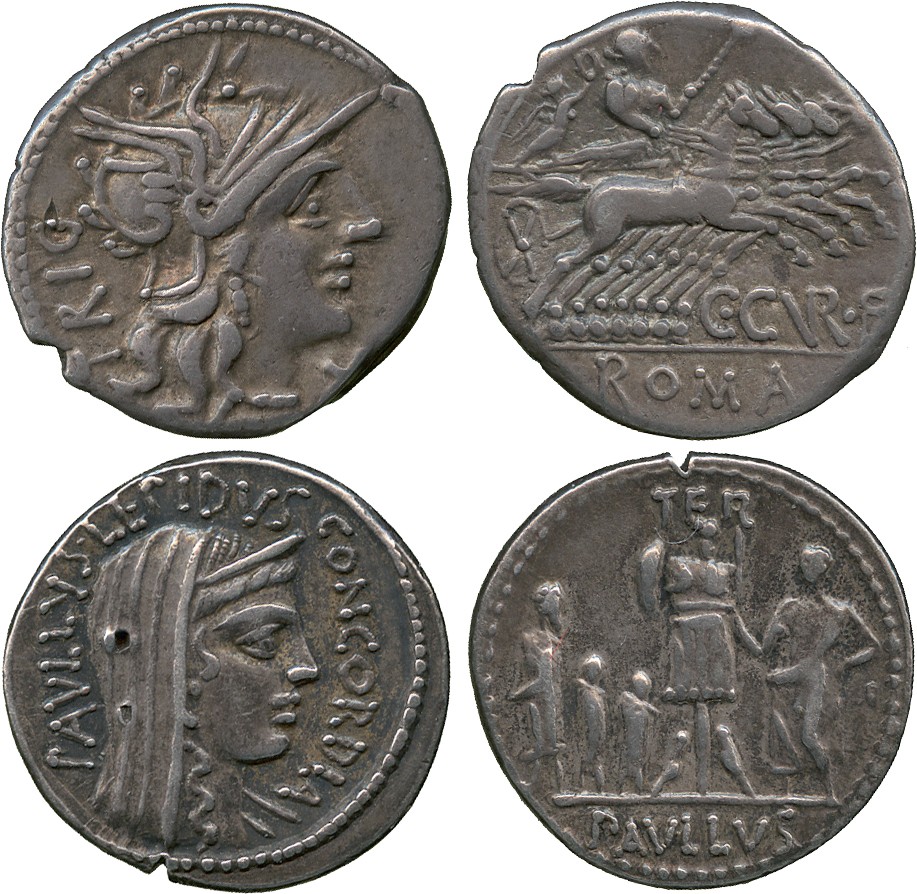}
        \caption{Multiple coins in one image}
    \end{subfigure}
    \hfill
    \begin{subfigure}[b]{0.4\textwidth}
        \centering
        \includegraphics[width=\textwidth]{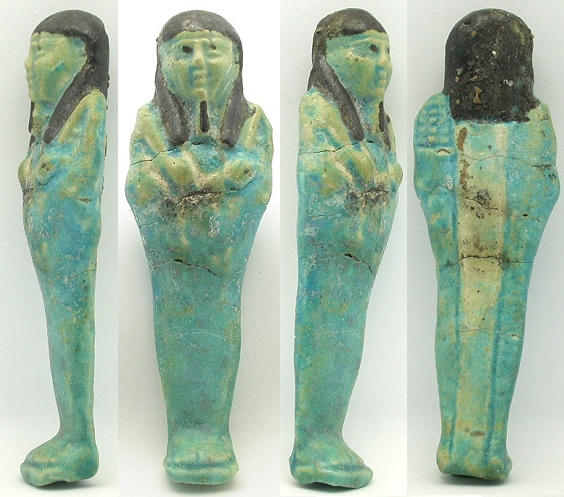}
        \caption{Not coins}
    \end{subfigure}
    \caption[Rejected coin images]{Examples of some of the images that were automatically rejected during preprocessing.}
    \label{fig:rejected-images}
\end{figure}

Since the primary aim of the preprocessing was to generate a data set that could be used for effectively training models to recognise semantic elements on a coin's reverse, it was desirable to identify and exclude any images that did not conform to the expected content and its presentation; examples include batch lots (i.e.\ lots containing multiple coins, often shown in a heap or otherwise overlapping, as shown in Figure~\ref{fig:rejected-images}) or artefacts other than coins.
In the case of an image showing multiple coins, even if the reverse sides could be isolated and an image per reverse side generated, the labels generated for all of them would be derived from one common image description, meaning that they would all have the same labels, regardless of their content, which would add noise to the training data.

The algorithm that was developed to preprocess each image worked as follows:

\renewcommand{\labelenumii}{\theenumii}
\renewcommand{\theenumii}{\theenumi.\arabic{enumii}.}

\begin{enumerate}
    \item Try to identify a background colour for the image by taking the mean colour of the pixels in the corners. If the difference between the colours in each corner is greater than some threshold, reject the image, as it does not have a consistent background colour, making image processing more awkward.
    \item Identify a bounding box for the bottom rightmost foreground object (which is assumed to be a coin's reverse side):
    \begin{enumerate}
        \item From the right of the image, find the first column whose pixels are not all of the background colour (within some tolerance). This is assumed to the right of the coin. If none is found, reject the image.
        \item Find the next column to the left that appears to be background-only. This is assumed to be just off the left of the coin. If none is found, the object must run to the edge of the image and may have been cropped, so reject it. If the distance between the right and the left is too small, go back to the previous step, but searching for a new `coin right' to the left of the current column.
        \item Repeat the previous two steps but performed vertically from the bottom instead, using rows instead of columns, to find the bottom and top of the coin. If these cannot be found, reject the image.
        \item Additionally, if the bounding box that was found significantly deviates from square (unlikely for a single coin) or it contains a significant percentage of background-coloured pixels (possibly indicative of a cluster of overlapping coins), repeat the search for a foreground object, starting to the left of the rejected one.
        \item Continue searching above the coin to find any additional foreground object that may be above it. If one is found, reject the image, as it does not conform to the expected layout.
    \end{enumerate}
    \item Identify a bounding box of the bottom rightmost foreground image to the left of the previous object found by repeating the previous step (and substeps) but starting from the column to the left of the previous object found. If an object is found, it is assumed to be a coin's obverse side; otherwise, reject the image.
    \item Continue searching for any additional foreground objects farther to the left of the last object found. If any are found, reject the image.
    \item Create a new image containing only the pixels contained in the first bounding box identified, which is assumed to contain the reverse side of one coin.
\end{enumerate}

\begin{figure}[h]
    \centering
    \includegraphics[width=10cm]{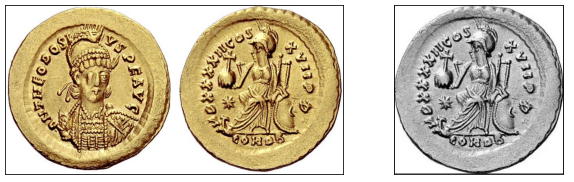}
    \caption[Processed coin image]{The algorithm successfully preprocessed the image on the left to the generate the one on the right.} 
    \label{fig:coin-success}
\end{figure}

\begin{figure}
    \centering
    \includegraphics[width=10cm]{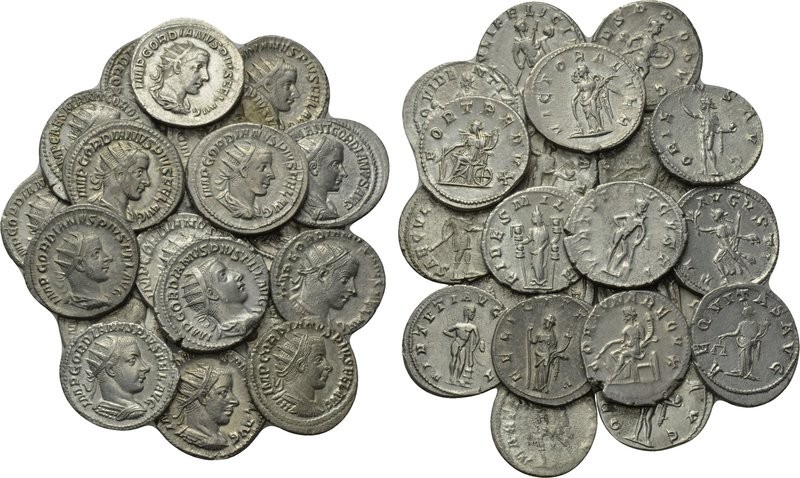}
    \caption[Bad coins image sample]{An example of an image that was not rejected during preprocessing but ideally would have been, as it includes multiple coins.}
    \label{fig:not-rejected}
\end{figure}

This algorithm makes some assumptions and is by no means perfect. In a manual inspection of a few thousand of the images that it generated, we found only a handful of images that did not feature just one single side of a coin, as desired; each of these featured a solid cluster of overlapping coins with a near-square bounding box (Figure~\ref{fig:not-rejected}) rather than a single coin. Based on the small number that were observed, these samples are believed to make up less than 0.2\% of the resultant data set, which is rare enough that they should not have had a significant impact on the performance of models trained using this data set. Therefore, we decided not to invest additional time to weed out the remaining unwanted samples.

\subsubsection{Image Labelling}

The automated process of labelling the images was quite simple and followed the method used by \cite{cooper_learning_2020}. For each of the semantic elements on coins chosen to be learnt by the models, each image was assigned a label based on whether its description contained any of a set of words related to the element. The set of search words consisted of the word for the element in English, French, Spanish, German and, in some cases, Latin, along with some synonyms and derived forms of the word (differing in grammatical number, case or gender). The translations of words were found by using Google Translate (https://translate.google.com). Some related search words (e.g. derived forms of verbs and German compound words containing a word of interest) were found by text searching for a word stem within a list of all the words that appeared in any of the descriptions. Finally, after some early models had been trained to recognise an element, we performed a manual inspection of some of the images that were classed as not containing the element and, in the case of false negatives (i.e. samples that contained the element but not classed as such), we identified further words indicative of the element, which we then added to the search words. The image labels were then regenerated and a new set of models trained using them.

The output of the data preprocessing was a smaller set of images (12.7\% of images were automatically rejected) showing the reverse sides of coins and a CSV file containing the labels for all of the images. The new images were saved in greyscale, as colour is not relevant to the shape of semantic elements depicted in coins, which the models were to be trained to recognise. This was done for efficiency in training the models, as the image data to be processed would be one third of the size (each pixel would have a single value rather than one for each of red, green and blue).

\subsection{ViT Training}

\subsubsection{Loading the Data Sets}

The data was split into training, validation and test sets (in a 64:16:20 ratio). The role of these different data sets is discussed in Section~\ref{sec:vit-train-models}. In the data set for training ViT models, each sample had multiple labels, corresponding to all of the semantic coin elements. It is best practice to try to ensure that the composition of the different data sets (training, validation and test) is as similar as possible, so that they do not have significant biases that may affect training and evaluation of the models. To this end, the data sets were \emph{stratified} across the different class labels, so that images of each of the semantic elements of interest were evenly split across the data sets. In the case of single-label samples, stratification is relatively straightforward. For multi-label samples, it is more complex, as the split of samples needs to satisfy the constraints of stratifying all of the class labels at the same time. To perform this split, we used an existing tool \cite{szymanski_scikitmultilearn_2019}, which uses an iterative method to allocate samples to data sets based on how much the sample's positive labels are `desired' by the given data set.

Since previous work had shown that ViT models needed large data sets to have performance comparable with the best CNN models \cite{dosovitskiy_image_2020}, all of the data available from the training set was used. The data sets naturally had a class imbalance with far more negative samples (not exhibiting a given concept) than positive samples. If a model were trained on a very unbalanced data set like this, it could end up performing with far lower accuracy on positive samples than negative samples, due to the relative rarity of positive samples in the training data. To counter this data set imbalance during training, larger weights were applied to positive samples in the ViT model's per-concept loss functions, such that each positive sample would have a larger effect during training.

\subsubsection{Creating the Models}

The ViT models that were trained in this work were initialized using weight parameters taken from a ViT model that was pretrained on ImageNet-21K. The ViT models were created using a library called PyTorch Image Models \cite{wightman_pytorch_2019}. There are a number of different sizes of ViT architectures. There was a choice of patch size (16x16 or 32x32), which determines the dimensions of the patches that each image is split into. A smaller patch size causes each image to be mapped to a larger number of patch embedding vectors, which means that the model receives more data for each image, generally enabling more accurate classifications. We opted to use the smaller patch size, as it offered significantly better accuracy \cite{dosovitskiy_image_2020}, albeit at a greater computational expense during training and use. ViT models come in different sizes (from `tiny' to `huge'), which ultimately affects the number of learnable parameters that they contain and the computational cost of using them. We opted to use a `large' variant, which had been pretrained on ImageNet-21K. The difference in accuracy between the `large' and `huge' variants was just a fraction of a percent \cite{dosovitskiy_image_2020} and so we decided to go for the less computationally expensive model with less than half as many parameters.

After a model with pretrained weights was loaded, the final layer in the model needed to be replaced. The final layer in the model (the `head') is a linear layer that maps a feature vector output by the Transformer encoder to values for each of the class labels (i.e. it computes $y=Ax + b$). The pretrained model comes with a head that is fine-tuned to classify the 1000 classes ImageNet. To make the model suitable for fine-tuning on the coins data set, the head was replaced with a new randomly initialized linear layer that mapped to the number of classes in the coins data set. 

Before the model was fine-tuned for classification of semantic elements on ancient coins, all of the layers in the model, apart from the final linear layer, were frozen so that training would not adjust their parameters. This is a common practice and is done so as to avoid undoing the learning that has already taken place during the pretraining of the model on millions of images. Intuitively, the model already has a general understanding of features that occur in images; we just want to train it to understand how those features relate to the domain-specific visual elements that we are interested in (horse, shield etc.).

Since we were freezing all layers except for the head, we were able to train a single model on all of the classes simultaneously without their interfering with one another. For a given image, all of the computation up to the final layer is the same for each of the classes. In the head, the parameters that are learnt are independent for each output class. Therefore, during fine-tuning, the parameters can be adjusted for each class independently. We opted to train a single model for all classes rather than one model per class for efficiency reasons: much of the computation during training would be the same for all classes.

\subsubsection{Training the Models}
\label{sec:vit-train-models}

To train the ViT models required an optimization algorithm and a loss function. For classification problems, the cross-entropy loss function is commonly used, as it provides a measure of the difference between two distributions (the predicted probabilities for each label and their true probabilities). This generalises to labels that may take on an arbitrary number of values. Since the labels in the coins data set are all binary (indicating whether a particular semantic element is present in the sample image), we used binary cross entropy as the loss function. Whereas the cross entropy loss function takes a predicted probability distribution over the possible values for each label as one of its inputs (the other input being the true labels), the binary cross entropy loss function takes a single value per label instead, indicating the probability that the label is positive (1) rather than negative (0). The formula for binary cross-entropy (BCE) loss for a predicted label $x$ and a true label $y$ is shown in Equation~\ref{eqn:bce}. For a true positive label, $y=1$ and the equation simplifies to $\log x$; for a true negative label, $y=0$ and the equation simplifies to $\log(1 - x)$.

\begin{equation}
    \text{BCE}(x, y) = y \log x + (1 - y) \log(1 - x)
    \label{eqn:bce}
\end{equation}

The optimization algorithm used for fine-tuning the ViT models was stochastic gradient descent (SGD) with momentum of 0.9, as was used by Dosovitskiy et al.\ \cite{dosovitskiy_image_2020}. SGD is a variant of gradient descent in which model parameters are adjusted after only processing a batch of samples rather than after processing the full data set. Each parameter is adjusted in proportion to the gradient of the loss with respect to that parameter as calculated over the batch. The gradients are automatically calculated by PyTorch's autodiff functionality \cite{pytorch_autodiff}, using the backpropagation algorithm, which depends on the chain rule for differentiation. Momentum is an optional technique that may be applied to SGD to help smooth out noise in the training data. Essentially, the optimizer applies a moving average across weight adjustments. It remembers between steps the previous adjustments it made to each weight and when calculating a new weight adjustment, it adds in a term proportional to the previous adjustment. In this way, the state of the optimizer can be thought of as having something analogous to physical momentum that can be accumulated over multiple updates, causing it to accelerate down slopes when the gradient is relatively unchanging, leading to faster convergence. Small fluctuations in the loss landscape due to noise in the batch's data (e.g. due to the use of the batch's gradient as a proxy for the gradient calculated over all possible samples) have a reduced impact on training when momentum is used.

One of the other important parameters to the optimizer is the learning rate. For too large a learning rate, training may not converge at all. For too small a learning rate, convergence can either take too long or lead to poor local minima (less likely when momentum is used). We experimented with learning rates between $10^{-2}$ and $10^{-5}$ and found that a learning rate of $10^{-4}$ gave results comparable to those of a learning rate of $10^{-5}$ but in a shorter training time.

\begin{figure}
    \centering
    \includegraphics{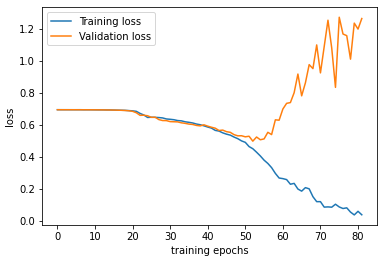}
    \caption[Learning curve]{A typical learning curve that exhibits overfitting beyond a point during training.}
    \label{fig:learning-curve}
\end{figure}

The models were optimized on the training set, while their performance on the validation set was monitored to see how well the model generalised. Typically, beyond a point during training, the performance of a model on the validation set will start to degrade (i.e. validation loss will start to increase as in Figure~\ref{fig:learning-curve}), which indicates that the model is starting to \emph{overfit}---optimising to the the specifics of the data included in the training set but to the detriment of performance on unseen data outside of the training set.

At the end of each training epoch (a full iteration over all of the training data), a checkpoint was saved containing the model's current parameters and performance statistics (loss, F1 scores and confusion matrices). These statistics were used to identify when to stop training the model, due to vastly diminishing returns or worsening performance. The best model saved during training could then be used for evaluation purposes.

\subsection{CNN Training}

\subsubsection{Loading the Data Sets}

The code for loading the data sets for training CNN models was almost identical to that which was used in the case of ViT models. The key difference was that the data sets only loaded the labels for a single semantic element, so all of the samples were single-label. This was done, because separate CNN models were trained for each semantic element of interest. In contrast to fine-tuning the head of a ViT model, when a CNN model is trained from scratch, all of the weights in the model are updated, and so a single model cannot be used for multi-label classifications without it having an impact on training. This meant that instead of using the iterative algorithm that was used for splitting up the data into training, validation and test sets for the ViT models, a more straightforward train/test split algorithm from Scikit-Learn \cite{pedregosa_scikitlearn_2011} could be applied instead.

Whereas the ViT data sets contained far more negative samples than positive samples, to create the CNN data sets, the negative samples were undersampled (i.e. not all were included), so that the resulting data sets contained an equal number of positive and negative samples, following the method of Cooper and Arandjelovi\'c \cite{cooper_learning_2020}. Since there were so few positive samples for `Hercules' and early results showed that a model trained on such a small, balanced `Hercules' data set had much worse performance than the models for other semantic elements, an updated model was trained on a data set in which the positive Hercules samples were oversampled, meaning that they appeared in the training data multiple times, and additional negative samples were included to result in a balanced data set. This allowed more negative samples to be included in the training data, while maintaining a balanced data set, resulting in a model with better performance.

\subsubsection{Creating the Models}

\begin{figure}[h]
    \centering
    \includegraphics[width=\textwidth]{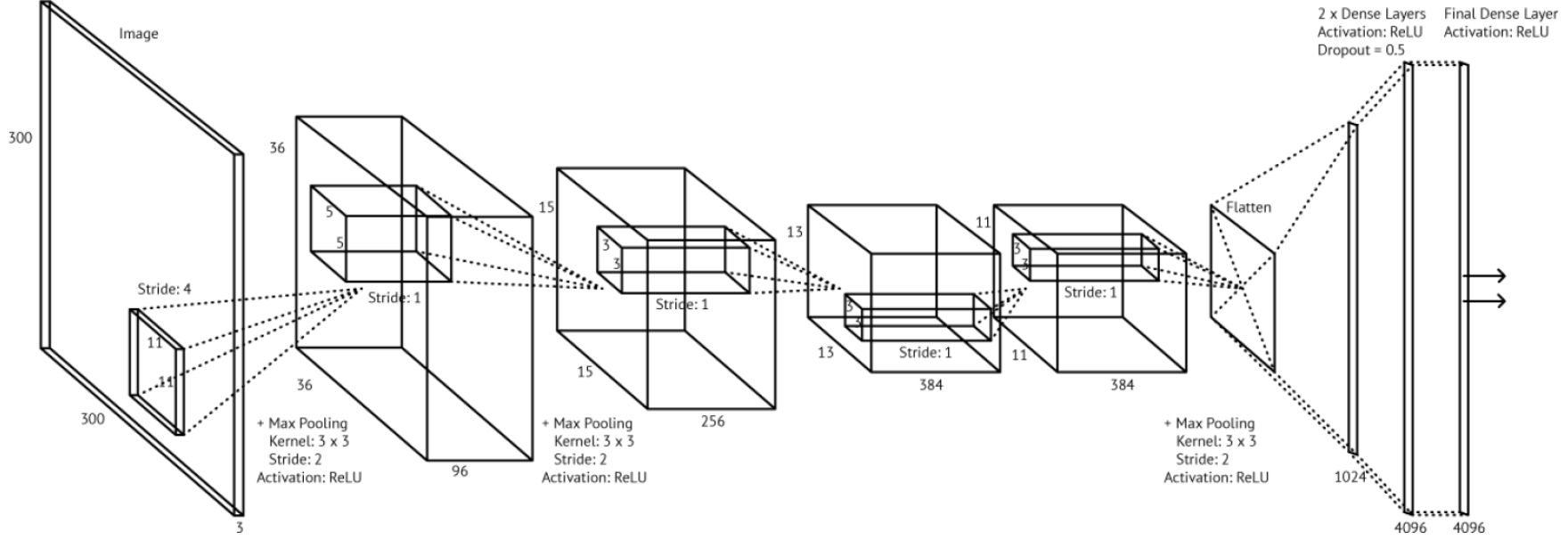}
    \caption[CNN network topology]{The CNN network topology used, as proposed by Cooper and Arandjelovi\'c \cite{cooper_learning_2020}.}
    \label{fig:cnn-topology}
\end{figure}

The CNN models that were trained were created out of PyTorch neural network modules that were composed following the network topology defined by Cooper and Arandjelovi\'c \cite{cooper_learning_2020} (Figure~\ref{fig:cnn-topology}). One issue with the model as it was defined was that it applied the ReLU activation function to the final layer of the network, which resulted in a model that during training tended to converge to a state of producing constant outputs (i.e. all ones or all zeros). Since the output of the final layer is passed to the loss function, which in this case had a domain of $\mathbb{R}$, there was no reason to apply ReLU to the final layer. As ReLU in the final layer appeared to be problematic for training, it was not applied there.

\subsubsection{Training the Models}

The code that was developed for training the CNN models was again very similar to that used for training ViT models. There were some small differences due to the fact that the samples for CNN training only had a single label per image. As the method described Cooper and Arandjelovi\'c \cite{cooper_learning_2020} was largely followed for CNN training, the cross-entropy loss function was used. The optimizer used by Cooper and Arandjelovi\'c \cite{cooper_learning_2020} was Adam, but our use of this in training was found to yield only models that output constant values. Inspection of the activations at each layer of the CNN showed that after only a few layers, the inputs to the ReLU function were mostly negative, meaning that the output for the layer became almost all zeros---a poorly understood phenomenon known as `dying ReLU' \cite{lu_dying_2020}. This behaviour was not observed when SGD with momentum was used as the optimizer instead. Since this was what was used for training the ViT models, SGD with momentum was used to train the CNNs (using a learning rate of $10^{-3}$ and momentum of 0.9).

Each CNN was only trained on a single semantic element and since they were much smaller models than the ViT models, training them was significantly quicker. Early stopping behaviour was added to the training loop, as described by Cooper and Arandjelovi\'c \cite{cooper_learning_2020}, such that if the model showed no performance improvement on the validation set for 30 training epochs, the training terminated.

\subsection{Text-Mining of Concepts}

The software for text-mining of concepts was very straightforward. It read in all of the descriptions of images in the data set, filtered out most punctuation symbols and a list of \emph{stop words} (unimportant common words, in this case, taken from NLTK \cite{bird_natural_2009}), and then generated a list of the remaining words in descending order of their frequency across descriptions in the data set. From this ranked list of words, three semantic elements (`seated', `standing' and `Hercules') were identified for further training, in addition to the five that were used by Cooper and Arandjelovi\'c~\cite{cooper_learning_2020} .





\section{Results and Discussion}\label{s:results}

This section provides a description of the data set used, an evaluation of the automated mining of concepts depicted on ancient coins, and an evaluation of the ViT and CNN models trained, comparing how they perform in the classification task of recognising common semantic elements depicted on ancient coins and making use of saliency maps to aid in understanding and explaining the models' behaviour. We discuss the experimental results in the context of previous research by Cooper et al.\ \cite{cooper_learning_2020}, which is the most similar prior research in this area.

\subsection{Data}\label{ss:data}
In this work, we made use of a data set, comprised of 100,000 images of ancient coins, provided by the Ancient Coins Search Engine (\url{https://www.acsearch.info/}) for research purposes, which has been used in a number of previous research efforts~\cite{zachariou2020visual,cooper2020learning}. This corpus consists of high-quality images obtained in rather controlled environments, usually with a uniform background, favourable lighting, natural coin alignment, etc. Whilst including a variety of non-Roman coins (Greek, Celtic, and Byzantine, among others), as well as Roman non-Imperial ones (namely Provincial and Republican), the Roman Imperial coins included span the entire time period of the Empire and cover most of the obverse figures depicted on them. 

\subsection{Evaluation of Automated Mining of Concepts}

Only a small amount of time was allocated to the development of software for the automated mining of coin concepts. Identified concepts were intended to be used in training models for automated coin analysis. The software developed counted words used in coins descriptions from the data set to identify the most common words, after a set of stop words had been excluded. This enabled the identification of additional concepts for use in model training. Three such concepts were selected: `seated', `standing' and `Hercules'. 

Since the coin descriptions from the data set referred to both sides of the coins, many of the concepts that were identified were more likely to refer to the obverse sides of coins (e.g. emperor names). Therefore, it still required some human input to filter this list down to produce a list of concepts that would be more suitable for model training. It would be ideal if the software were to be able to identify from the descriptions only the parts that were relevant to the reverse side, excluding descriptions of the obverse side or references to authoritative texts on numismatics (e.g. an author with the surname `Rider', who could be mistaken for a horse rider). This would be a somewhat complex language problem in itself and so is left for future work using this data set.

\subsection{Evaluation of Trained Models}

The experimental results for the trained models are shown in Table~\ref{tbl:summary}. To gain a qualitative understanding of the performance of the models, we manually inspected a subset of samples for each concept (e.g. `cornucopia') and each combination of actual and predicted labels (i.e. true positives, false positives, true negatives and false negatives). Additionally, we generated saliency maps for a subset of samples to aid in understanding model behaviour. The results and findings for each of the ancient coin concepts are discussed in turn.

\subsubsection{Results}

\newcommand{\rothead}[1]{\multicolumn{1}{p{0.8cm}}{\raggedleft\rotatebox{45}{\textbf{#1}}}}
\newcommand{\best}[1]{\textbf{#1}}

\begin{table}[h]
    \centering
    \begin{tabular}{l l *{8}{>{\centering\arraybackslash}m{0.8cm}}}
     & & \rothead{Cornucopia} & \rothead{Eagle} & \rothead{Horse} & \rothead{Patera} & \rothead{Shield} & \rothead{Standing} & \rothead{Seated} & \rothead{Hercules}\\
        \hline
        \multirow{11}{*}{\rotatebox{90}{ViT}}
         & Number of epochs        & 15          & 15          & 13          & 17          & 19          & 10          & 15          & 13\\
         & Training time / mins    & 2679        & 2679        & 2322        & 3036        & 3393        & 1786        & 2679        & 2322\\
         \cline{2-10}
         & Training accuracy       & 0.83        & 0.79        & 0.83        & 0.84        & 0.75        & 0.77        & 0.83        & 0.89\\
         \cline{2-10}
         & Validation accuracy     & 0.81        & 0.78        & 0.82        & 0.81        & 0.74        & 0.76        & 0.82        & 0.80\\
         & Validation precision    & 0.82        & 0.78        & 0.82        & 0.75        & 0.71        & 0.72        & 0.89        & 0.81\\
         & Validation recall       & 0.81        & 0.78        & 0.81        & 0.92        & 0.83        & 0.87        & 0.73        & 0.79\\
         & Validation F1           & 0.81        & 0.78        & 0.81        & 0.83        & 0.76        & 0.79        & 0.80        & 0.80\\
         \cline{2-10}
         & \textbf{Test accuracy}  & \best{0.80} & \best{0.77} & \best{0.81} & \best{0.83} & \best{0.72} & \best{0.76} & \best{0.82} & \best{0.83}\\
         & \textbf{Test precision} & \best{0.81} & \best{0.77} & \best{0.80} & 0.77        & \best{0.69} & \best{0.71} & \best{0.88} & \best{0.82}\\
         & \textbf{Test recall}    & 0.78        & \best{0.77} & \best{0.82} & \best{0.93} & \best{0.81} & \best{0.87} & 0.73        & \best{0.85}\\
         & \textbf{Test F1}        & \best{0.79} & \best{0.77} & \best{0.81} & \best{0.84} & \best{0.74} & \best{0.78} & \best{0.80} & \best{0.83}\\
        \hline
        \multirow{10}{*}{\rotatebox{90}{CNN}}
         & Number of epochs        & 53          & 70          & 41          & 75          & 57   & 15   & 20   & 238 \\
         & Training time / mins    & 18          & 26          & 17          & 15          & 40   & 20   & 11   & 26  \\
         \cline{2-10}
         & Training accuracy       & 0.81        & 0.77        & 0.79        & 0.87        & 0.68 & 0.76 & 0.84 & 0.67\\
         \cline{2-10}
         & Validation accuracy     & 0.76        & 0.69        & 0.76        & 0.83        & 0.62 & 0.73 & 0.79 & 0.67\\
         & Validation precision    & 0.71        & 0.73        & 0.77        & 0.86        & 0.67 & 0.71 & 0.82 & 0.65\\
         & Validation recall       & 0.86        & 0.60        & 0.72        & 0.78        & 0.47 & 0.78 & 0.74 & 0.73\\
         & Validation F1           & 0.78        & 0.66        & 0.75        & 0.82        & 0.55 & 0.74 & 0.77 & 0.69\\
         \cline{2-10}
         & \textbf{Test accuracy}  & 0.74        & 0.71        & 0.74        & 0.82        & 0.62 & 0.72 & 0.78 & 0.67\\
         & \textbf{Test precision} & 0.71        & \best{0.77} & 0.77        & \best{0.84} & 0.67 & 0.70 & 0.80 & 0.65\\
         & \textbf{Test recall}    & \best{0.82} & 0.59        & 0.70        & 0.79        & 0.46 & 0.77 & \best{0.74} & 0.74\\
         & \textbf{Test F1}        & 0.76        & 0.67        & 0.73        & 0.81        & 0.54 & 0.74 & 0.77 & 0.69\\
         \hline
    \end{tabular}
    \caption[Experimental results summary]{Summary of experimental results for the ViT and CNN models trained.}
    \label{tbl:summary}
\end{table}

The ViT and CNN models trained on each of the ancient coin semantic elements were tested on previously unseen test sets of samples, containing equal numbers of positive and negative samples per semantic element trained upon. The summary statistics for each model and associated semantic element are shown in Table~\ref{tbl:summary}.

As previously discussed, a single ViT model was trained on all of the semantic elements at once. The ViT training times are therefore larger than they would have been had a model per semantic element been trained. Even if the overall ViT model training time were to be divided between the different semantic elements, the training time would still far exceed the training times for the CNN models. A large part of this is down to the relative model sizes: the ViT model was far larger than the CNN models, having ~304M parameters rather than ~37M. A further difference was that the ViT model was trained on a larger number of negative samples, which meant that each training epoch covered more data.

In general, the statistics for the validation set and test set for a given model and semantic element are similar. This indicates that the validation and test data sets had a similar composition, which is required during training so that the performance on the validation set can be used as a predictor of performance on unseen data in the test set.

For each test statistic in Table~\ref{tbl:summary}, the higher value per column is highlighted in bold, enabling comparison between the ViT and CNN models. The test statistics for the ViT model are higher in all but a few cases, demonstrating generally superior performance for the ViT model over the CNN model.

\subsubsection{Hardware}

The computer hardware that we used for training and testing the performance of the ViT and CNN models consisted of an Asus H110M-Plus motherboard with an Intel Core i5 6500 3.2GHz Quad Core CPU, 32GB RAM and an NVidia GTX 1060 6GB GPU.

\subsubsection{Saliency Mapping}

As discussed in Section~\ref{sec:visual-explanations}, visual explanations, such as those provided by the saliency mapping tool HiPe \cite{cooper_believe_2022}, can be useful in helping one to understand how a deep learning model is behaving. In evaluating the models trained for this work, we made use of saliency maps, generated by HiPe. 

When we tried to generate saliency maps for the trained ViT model, we found that the GPU of the machine that we were using would consistently run out of memory before a map could be generated at higher resolutions. To resolve this issue, we merged two sequential loops in HiPe's core algorithm (which produced and consumed intermediate results respectively) into a single loop, so that the memory used for intermediate results could be freed at the end of each iteration of the merged loop.

For our purposes, we found that higher resolution saliency maps offered diminishing returns in the insight they provided to us. Therefore, we decided ultimately to use a fixed maximum depth so that we could generate saliency maps more efficiently and, thus, produce and analyse more of them.

\subsubsection{Cornucopia}

The ViT model had lower recall on the test set than the CNN model but overall it still had higher accuracy, due to its much higher precision than the CNN model. Both models often mistakenly classified as positive (i.e. as featuring a cornucopia) images of figures holding objects other than a cornucopia.

Although much less common, some coins featured a very large cornucopia that occupied most of the reverse. The CNN model generally failed to recognise these as being cornucopias, whereas the ViT model had slightly more success in this regard, demonstrating a more generalised understanding of this concept.

For both models, the set of `false positives' contained many images that did, in fact, feature a cornucopia, demonstrating that the labels, based on coin descriptions, were noisy and often inaccurate.

\begin{figure}[h]
    \begin{minipage}{0.47\textwidth}
    \includegraphics[width=\textwidth]{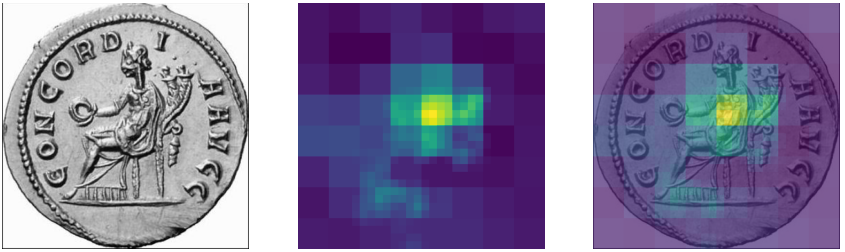}
    \includegraphics[width=\textwidth]{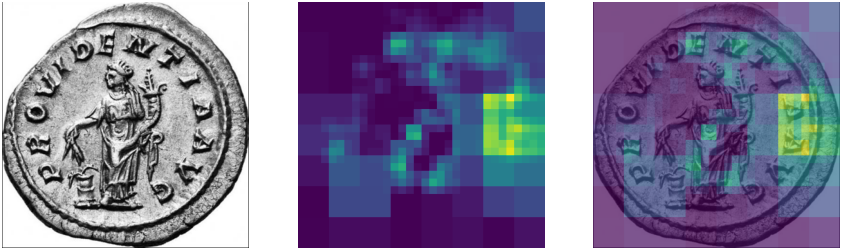}
    \end{minipage}
    \hspace{0.06\textwidth}
    \begin{minipage}{0.47\textwidth}
    \includegraphics[width=\textwidth]{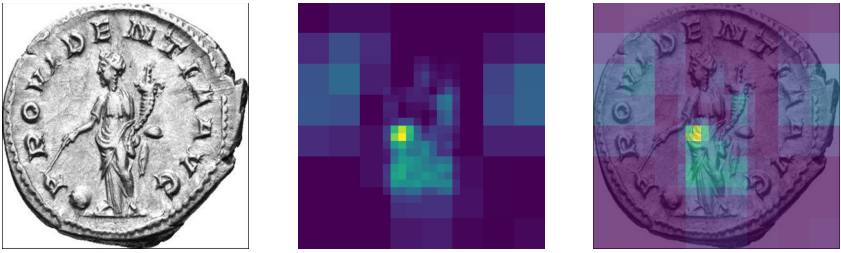}
    \includegraphics[width=\textwidth]{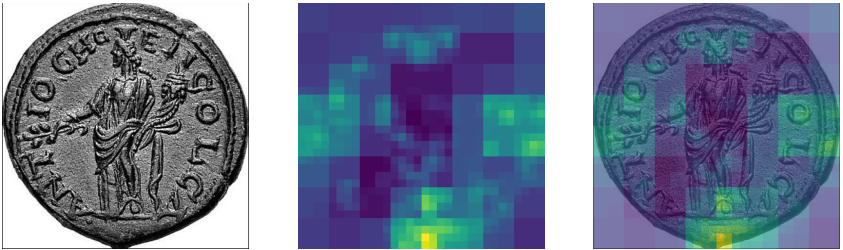}
    \end{minipage}
    \caption[ViT cornucopia saliency maps]{The `cornucopia' ViT model often had different regions of saliency for similar images.}
    \label{fig:vit-cornucopia}
\end{figure}

Saliency maps for the ViT model showed great diversity in what regions were highlighted, even for remarkably similar images (Figure~\ref{fig:vit-cornucopia}). The most salient regions rarely seemed to include the cornucopia itself, suggesting that the model learnt other predictors for the presence of a cornucopia. In contrast, the CNN model appeared to have learnt to focus on the same position for every image: the top-right, where a cornucopia is most commonly depicted.

\begin{figure}[h]
    \begin{minipage}{0.47\textwidth}
    \includegraphics[width=\textwidth]{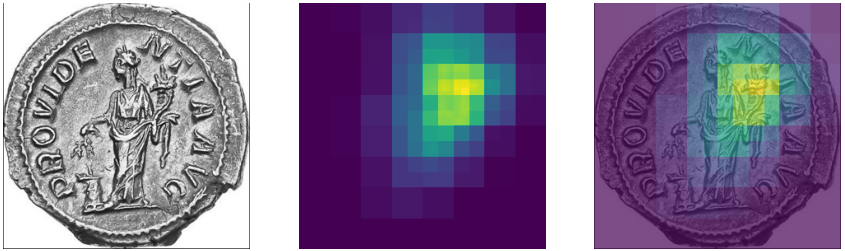}
    \includegraphics[width=\textwidth]{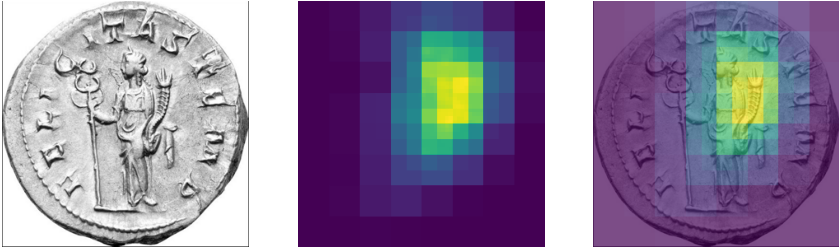}
    \end{minipage}
    \hspace{0.06\textwidth}
    \begin{minipage}{0.47\textwidth}
    \includegraphics[width=\textwidth]{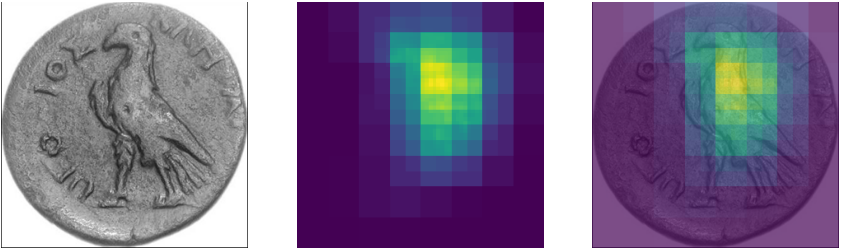}
    \includegraphics[width=\textwidth]{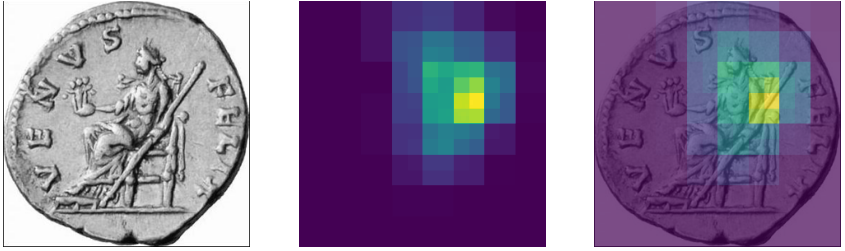}
    \end{minipage}
    \caption[CNN cornucopia saliency maps]{The `cornucopia' CNN model appeared to only focus on the top-right of each image. The images on the left are true positives; those on the right are false positives.}
    \label{fig:cnn-cornucopia}
\end{figure}

\subsubsection{Eagle}

From Table~\ref{tbl:summary}, it can be seen that the ViT model generally outperformed the CNN model for `eagle'. The CNN model tied with the ViT model for precision but its very low recall score compared poorly to the ViT model.

\begin{figure}[h]
    \begin{minipage}{0.47\textwidth}
    \includegraphics[width=\textwidth]{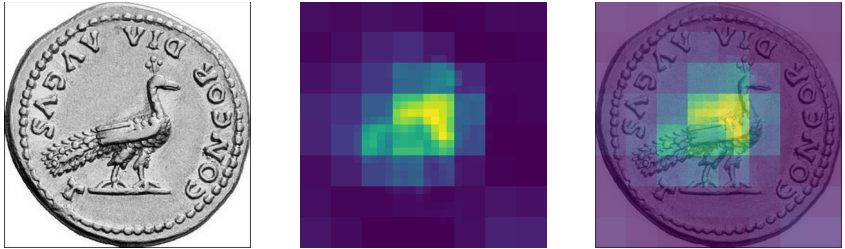}
    \includegraphics[width=\textwidth]{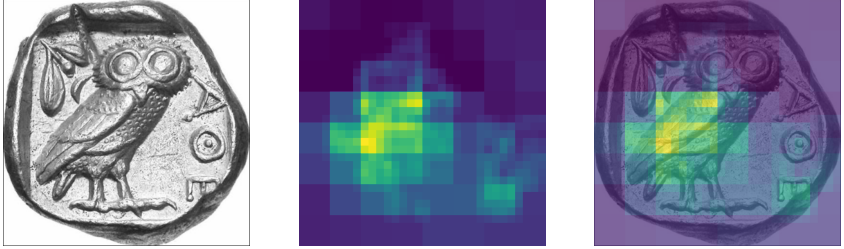}
    \end{minipage}
    \hspace{0.06\textwidth}
    \begin{minipage}{0.47\textwidth}
    \includegraphics[width=\textwidth]{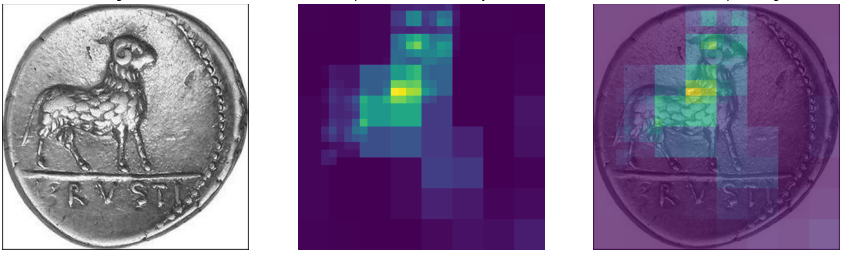}
    \includegraphics[width=\textwidth]{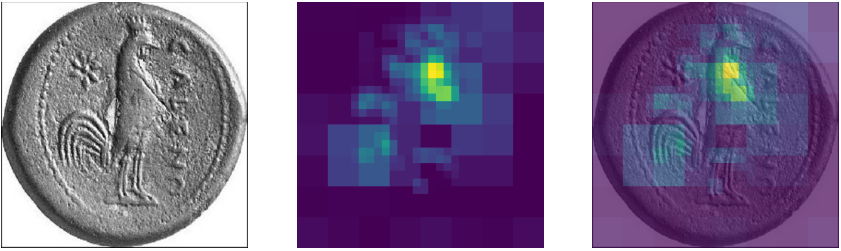}
    \end{minipage}
    \caption[ViT eagle saliency maps]{Examples of false positives for the `eagle' ViT model that feature patterns that could be mistaken for feathers (including on a goat).}
    \label{fig:vit-eagle-fp}
\end{figure}

As was the case for cornucopia detection, the saliency maps for the ViT model seem to be remarkably variable, so it is hard to identify what visual cues the model has learnt to associate with eagles. Present among the false positives for the ViT model are some other birds and animals, for which the focus appears to be on patterned regions, which could be mistaken for eagle feathers (Figure~\ref{fig:vit-eagle-fp}).

\begin{figure}[h]
    \centering
   \includegraphics[width=0.5\textwidth]{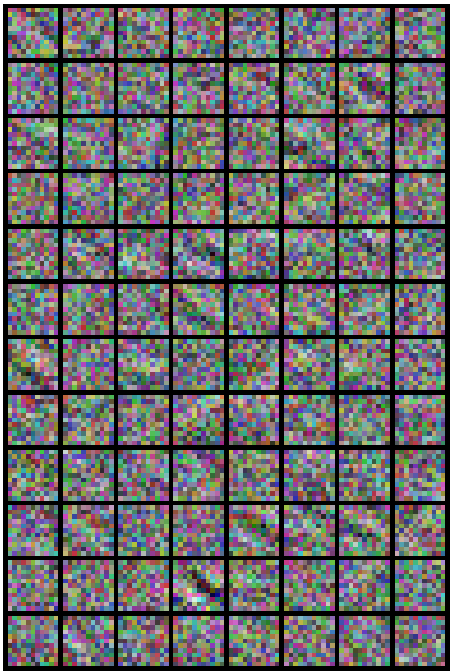}
   \caption[CNN eagle convolutional filter]{The most prominent patterns in the filters for the first layer of the `eagle' CNN model contained diagonal lines from the top-left to bottom-right, suited to identifying feathers at the same angle.}
   \label{fig:cnn-eagle-filter}
\end{figure}

\begin{figure}
    \begin{minipage}{0.47\textwidth}
    \includegraphics[width=\textwidth]{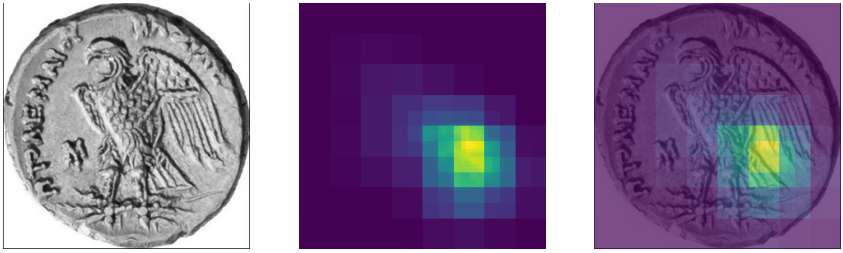}
    \includegraphics[width=\textwidth]{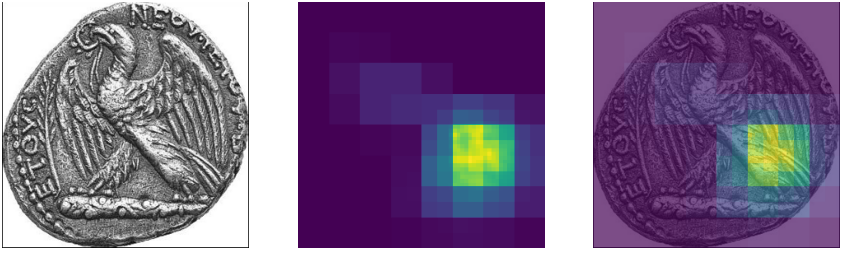}
    \end{minipage}
    \hspace{0.06\textwidth}
    \begin{minipage}{0.47\textwidth}
    \includegraphics[width=\textwidth]{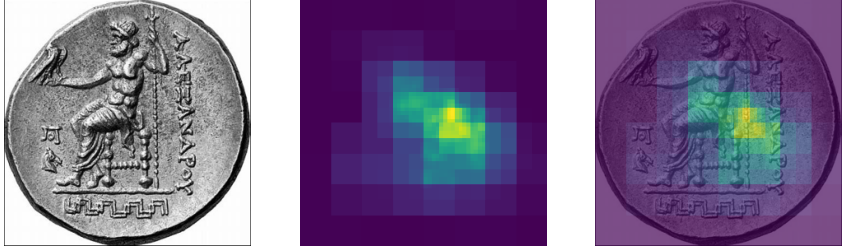}
    \includegraphics[width=\textwidth]{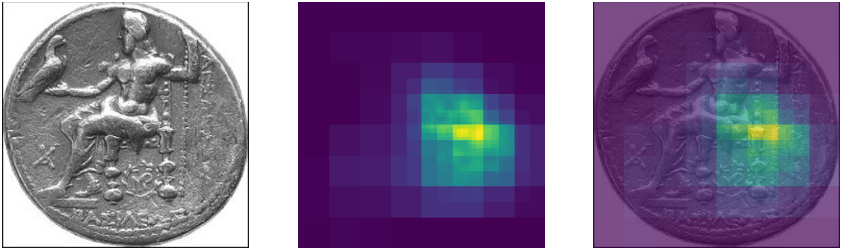}
    \end{minipage}
    \caption[CNN eagle saliency maps]{True positives for the `eagle' CNN model. Left: the salient region in the bottom left features feathers. Right: the salient region does not include the eagle.}
    \label{fig:cnn-eagle-tp}
\end{figure}

The CNN eagle model mostly focused on the lower right, where the wing or tail feathers of a full-size eagle could often be found for positive eagle samples, with lines on the top-left to bottom-right diagonal (Figure~\ref{fig:cnn-eagle-tp}). A focus on such diagonal lines was evident in the learnt filters for the first convolutional layer of the CNN (Figure~\ref{fig:cnn-eagle-filter}). Eagles facing to the right, which were slightly rarer, appeared to have been missed by the model more often, likely due to the focus on the bottom right and diagonal lines towards it.

A substantial fraction of positive eagle samples featured Jupiter/Zeus sat on a chair holding an eagle in an arm outstretched to the left of the coin. The salient regions for such images did not include the eagle but instead the back of the chair, where a sceptre ran down (see right of Figure~\ref{fig:cnn-eagle-tp}). This demonstrates how weak labels that only apply to the whole image may result in a model that recognises different visual elements correlated with the label than the semantic element of interest.

\subsubsection{Horse}

\begin{figure}[h]
    \begin{minipage}{0.47\textwidth}
    \includegraphics[width=\textwidth]{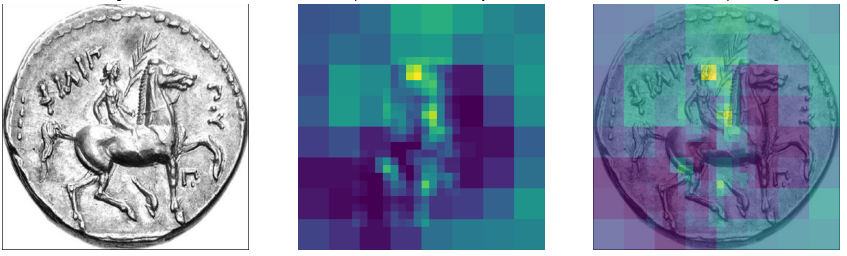}
    \includegraphics[width=\textwidth]{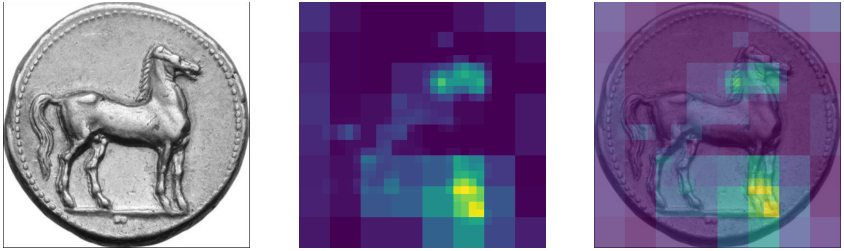}
    \end{minipage}
    \hspace{0.06\textwidth}
    \begin{minipage}{0.47\textwidth}
    \includegraphics[width=\textwidth]{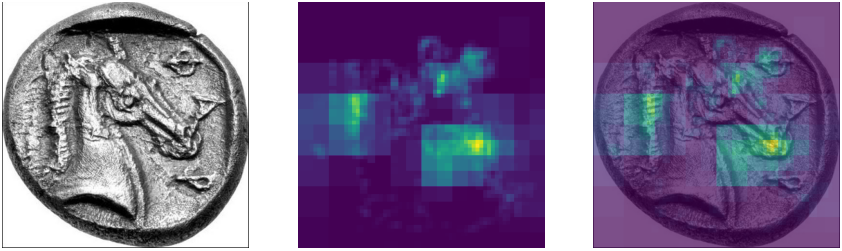}
    \includegraphics[width=\textwidth]{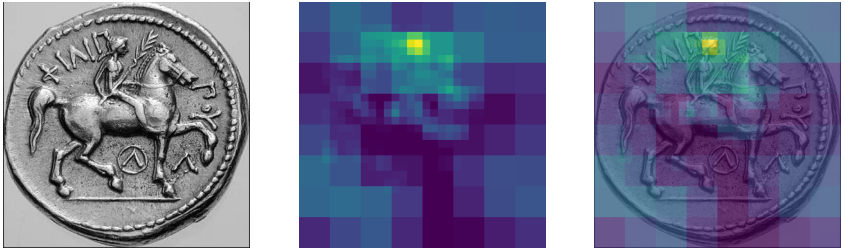}
    \end{minipage}
    \caption[ViT horse saliency maps]{The salient regions of `horse' for the ViT model were more varied and dispersed, suggesting that the model had learnt a wider variety of visual cues for horses than the CNN model had.}
    \label{fig:vit-horse}
\end{figure}

The ViT model for horses outperformed the CNN model in terms of the test statistics recorded. Again, the ViT model appeared to have quite variable salient regions, which were not very easy to interpret (Figure~{\ref{fig:vit-horse}}).

\begin{figure}[h]
    \begin{minipage}{0.47\textwidth}
    \includegraphics[width=\textwidth]{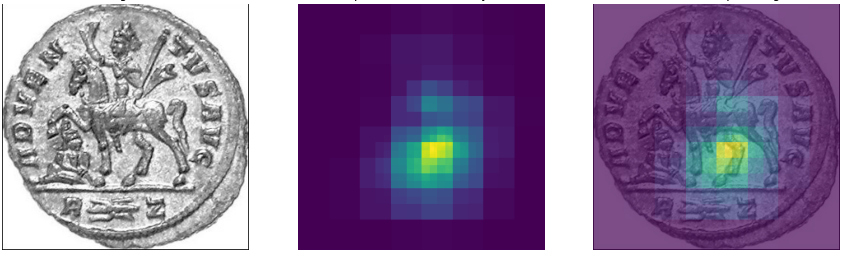}
    \includegraphics[width=\textwidth]{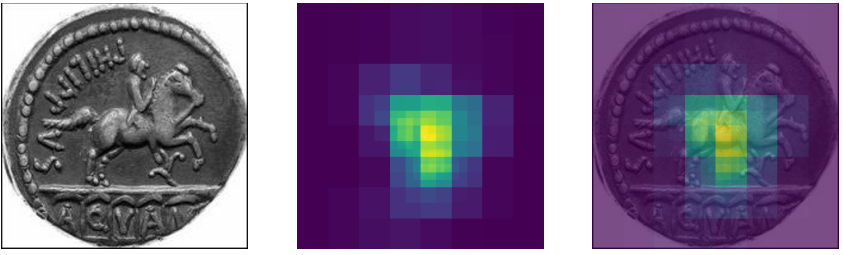}
    \end{minipage}
    \hspace{0.06\textwidth}
    \begin{minipage}{0.47\textwidth}
    \includegraphics[width=\textwidth]{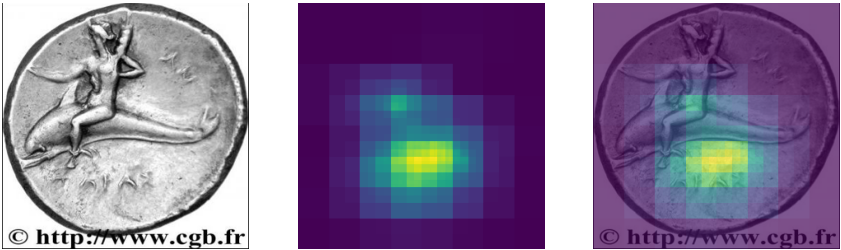}
    \includegraphics[width=\textwidth]{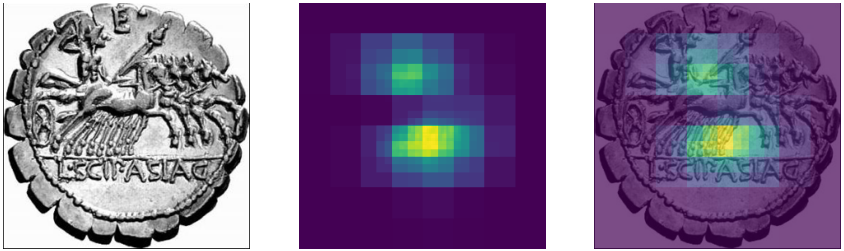}
    \end{minipage}
    \caption[CNN horse saliency maps]{The `horse' CNN model often focused just below the centre, which might be horse legs, the underside of a horse or a dolphin.}
    \label{fig:cnn-horse}
\end{figure}

The sample labels appeared to be quite noisy. A lot of `false positive' samples actually contained a horse, meaning that the labels were wrong. Some of these samples included `quadriga' or `biga' in their descriptions, which would imply the presence of a horse; however, these terms were not included as search words in the labelling process and so the samples were labelled as negative for horses. A number of dolphin riders featured among the `true positives', due to their descriptions mentioning a horse on the obverse side, rather than the reverse (Figure~\ref{fig:cnn-horse}).

\subsubsection{Patera}

The ViT model had very similar accuracy on pateras to the CNN model. The false positives for both models nearly always featured a person, often with an arm outstretched to the left of the coin, holding an object other than a patera. The true positives mostly matched this form, but the figure actually was holding a patera instead.

\begin{figure}[h]
    \begin{minipage}{0.47\textwidth}
    \includegraphics[width=\textwidth]{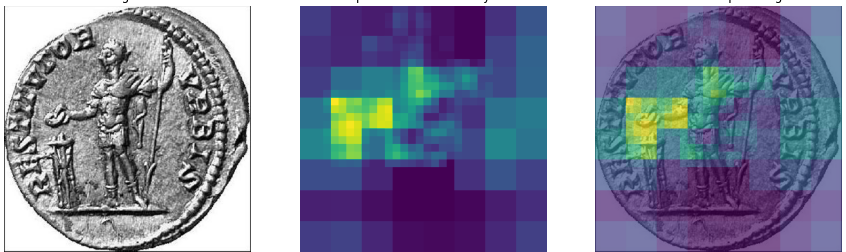}
    \includegraphics[width=\textwidth]{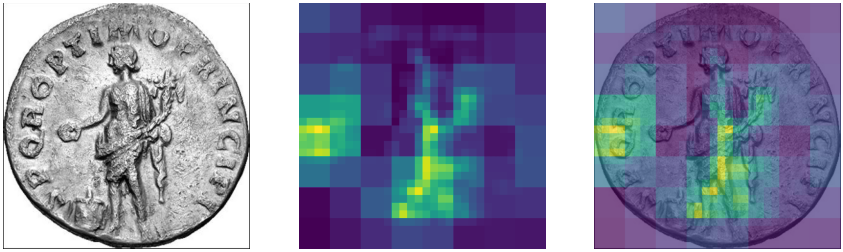}
    \end{minipage}
    \hspace{0.06\textwidth}
    \begin{minipage}{0.47\textwidth}
    \includegraphics[width=\textwidth]{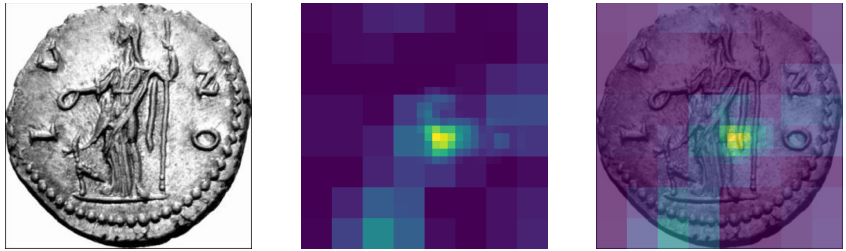}
    \includegraphics[width=\textwidth]{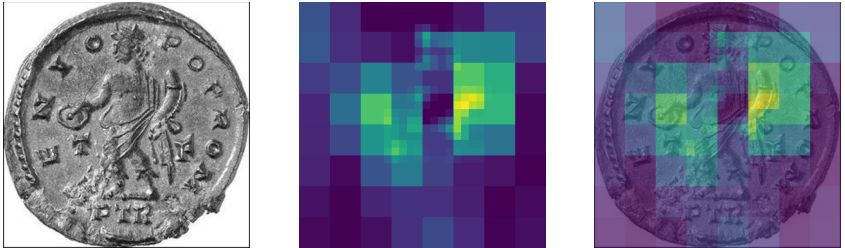}
    \end{minipage}
    \caption[ViT patera saliency maps]{The salient regions of the ViT model's true positives for patera were varied, but often included the patera.}
    \label{fig:vit-patera}
\end{figure}

For the ViT model, the salient regions were again quite variable, but often included the patera (Figure~\ref{fig:vit-patera}). 

\begin{figure}[h]
    \begin{minipage}{0.47\textwidth}
    \includegraphics[width=\textwidth]{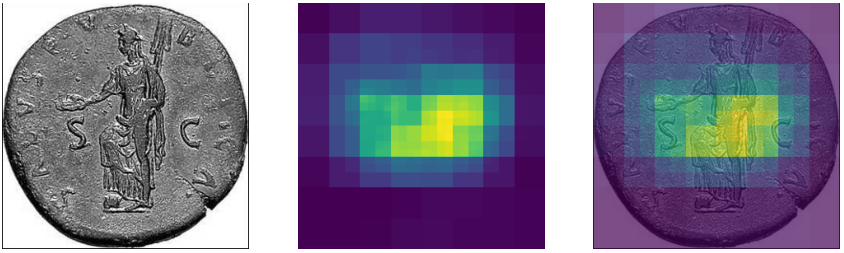}
    \includegraphics[width=\textwidth]{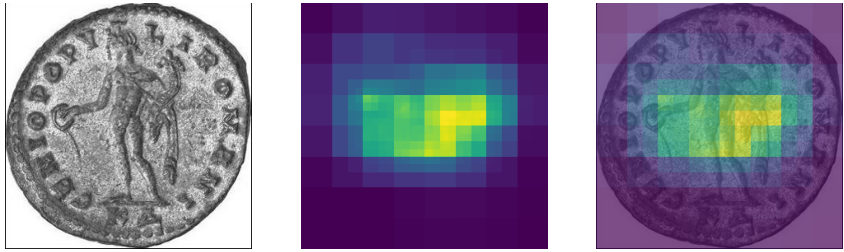}
    \end{minipage}
    \hspace{0.06\textwidth}
    \begin{minipage}{0.47\textwidth}
    \includegraphics[width=\textwidth]{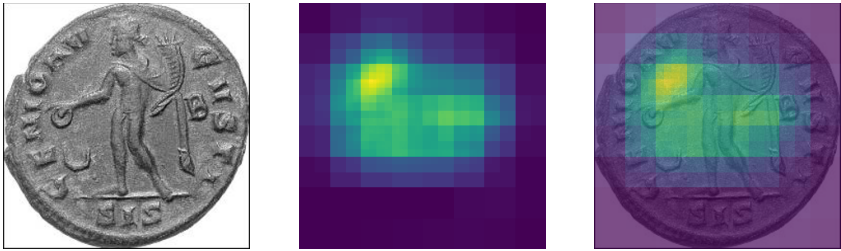}
    \includegraphics[width=\textwidth]{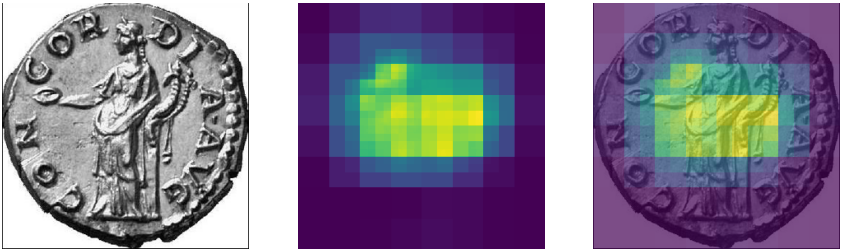}
    \end{minipage}
    \caption[CNN patera saliency maps]{The `patera' CNN model was once again more consistently focused on one position.}
    \label{fig:cnn-patera}
\end{figure}

In contrast, the salient regions for the CNN model almost run across a central horizontal strip that would span the middle of a person holding a patera. It also appears that a small empty region just above where the patera is held is "salient", more so than the region containing the patera itself. One possibility is that the model is sensitive to whether this region is empty. Many objects featured being held on coins are larger than a patera (e.g. spears or eagles) and so would fill this space.

\subsubsection{Shield}

\begin{figure}[h]
    \begin{minipage}{0.47\textwidth}
    \includegraphics[width=\textwidth]{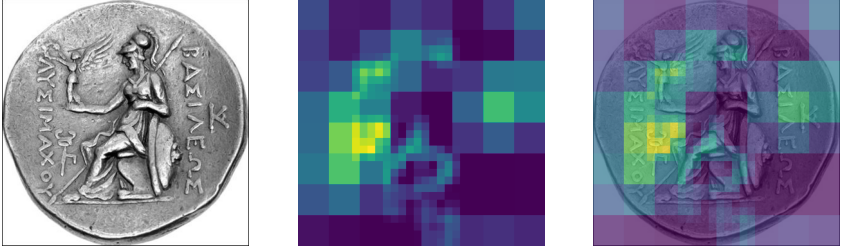}
    \includegraphics[width=\textwidth]{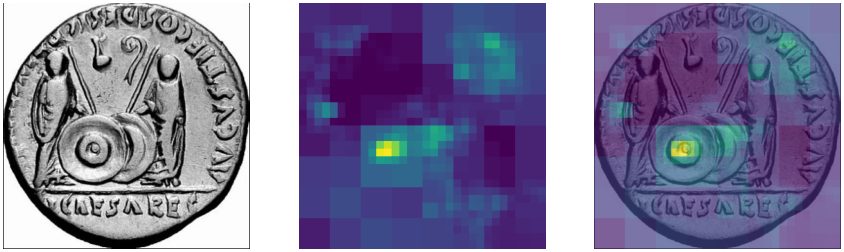}
    \end{minipage}
    \hspace{0.06\textwidth}
    \begin{minipage}{0.47\textwidth}
    \includegraphics[width=\textwidth]{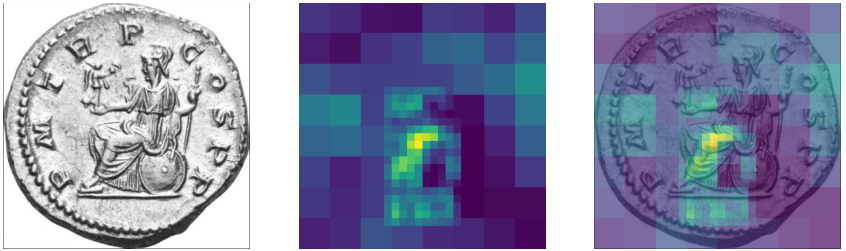}
    \includegraphics[width=\textwidth]{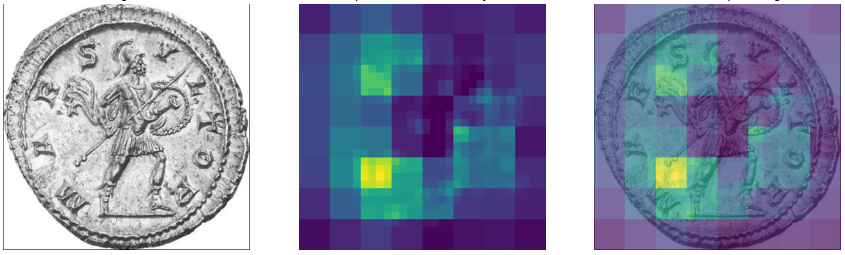}
    \end{minipage}
    \caption[ViT shield saliency maps]{The ViT model's salient regions for `shield' varied considerably and rarely included the shield itself.}
    \label{fig:vit-shield}
\end{figure}

\begin{figure}[h]
    \begin{minipage}{0.47\textwidth}
    \includegraphics[width=\textwidth]{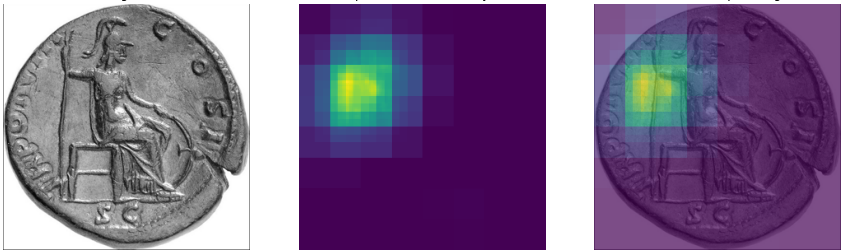}
    \includegraphics[width=\textwidth]{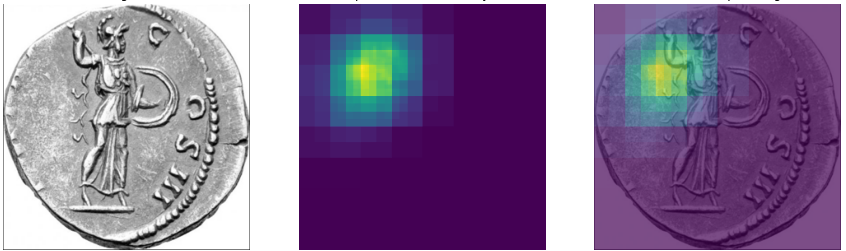}
    \end{minipage}
    \hspace{0.06\textwidth}
    \begin{minipage}{0.47\textwidth}
    \includegraphics[width=\textwidth]{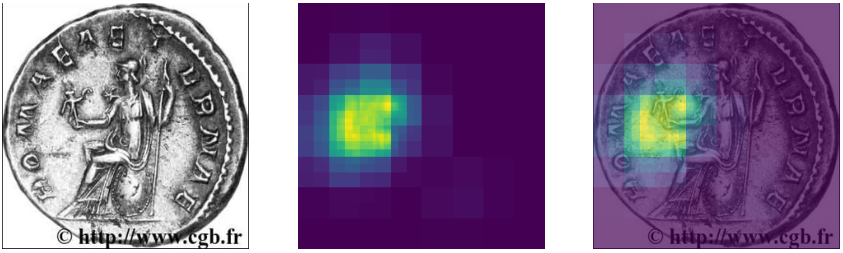}
    \includegraphics[width=\textwidth]{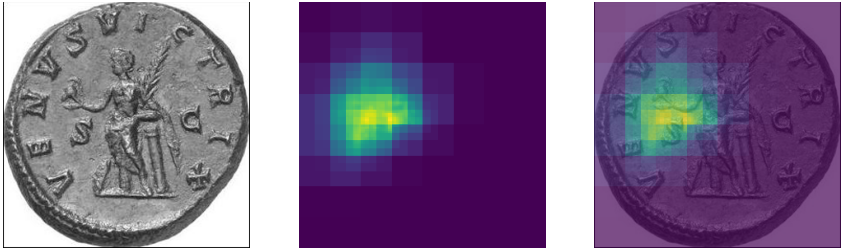}
    \end{minipage}
    \caption[CNN shield saliency maps]{The `shield' CNN model almost exclusively focused on the top-left, which, among positive samples, often included a bent arm holding a spear, or a cross carried by an angel. The latter appeared to be a common mislabelling, due to a shield on the obverse of such coins. }
    \label{fig:cnn-shield}
\end{figure}

The ViT model (Figure~\ref{fig:vit-shield}) outperformed the CNN model for shield detection (Figure~\ref{fig:cnn-shield}), but both models had the lowest accuracy among the semantic elements considered, despite requiring the longest training time, demonstrating the difficulty of learning this particular concept. This is likely due to the high variability among depictions of shields, as noted by Cooper and Arandjelovi\'c \cite{cooper_learning_2020}.

\subsubsection{Standing}

\begin{figure}[h]
    \begin{minipage}{0.47\textwidth}
    \includegraphics[width=\textwidth]{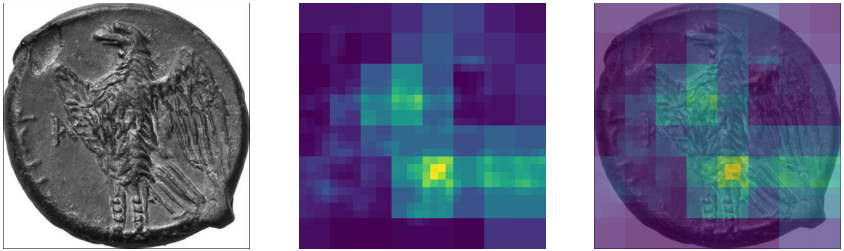}
    \includegraphics[width=\textwidth]{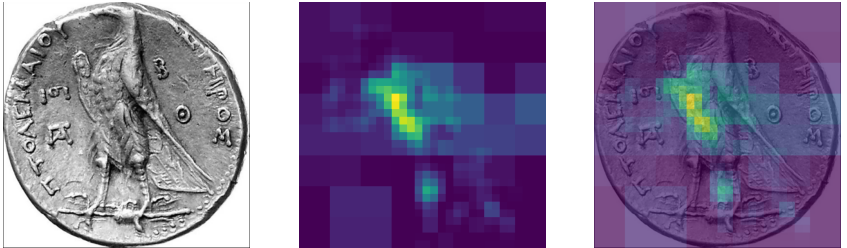}
    \end{minipage}
    \hspace{0.06\textwidth}
    \begin{minipage}{0.47\textwidth}
    \includegraphics[width=\textwidth]{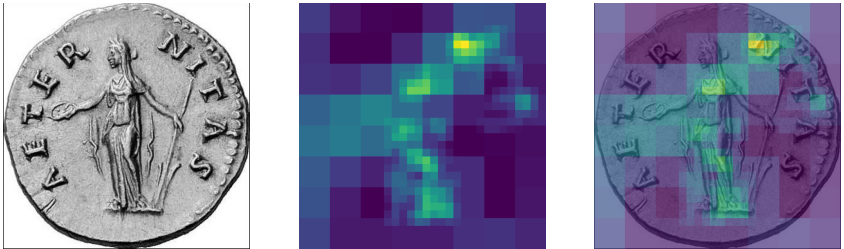}
    \includegraphics[width=\textwidth]{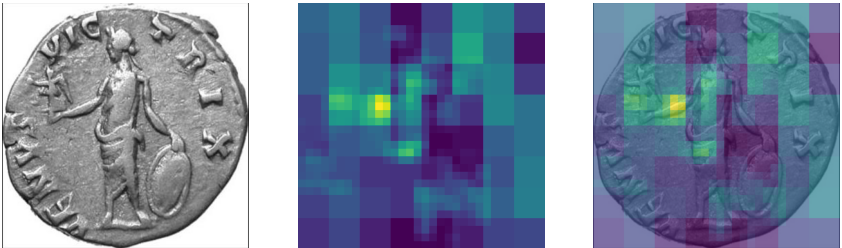}
    \end{minipage}
    \caption[ViT standing saliency maps]{True positive examples of `standing' for the ViT model.}
    \label{fig:vit-standing}
\end{figure}

The size of the data set used for `standing' was the largest among the semantic elements studied, which one might suspect would lead to better performance. Both the ViT and CNN models had relatively high recall but then low precision that brought down their overall accuracy (which was slightly higher for the ViT model). The reason for this is probably down to noisy labelling of samples. Many `false positive' samples actually featured standing, which meant that the models had correctly recognised standing but the coin descriptions did not explicitly mention standing. Very few `false negatives' appeared to show a person standing, but many contained standing animals, suggesting that the models had not learnt to recognise standing animals as well as standing humans. The ViT model managed to correctly classify some eagles, but few other animals (Figure~\ref{fig:vit-standing}).

\begin{figure}[h]
    \begin{minipage}{0.47\textwidth}
    \includegraphics[width=\textwidth]{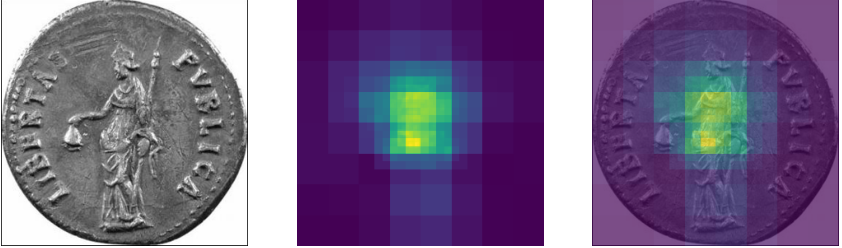}
    \includegraphics[width=\textwidth]{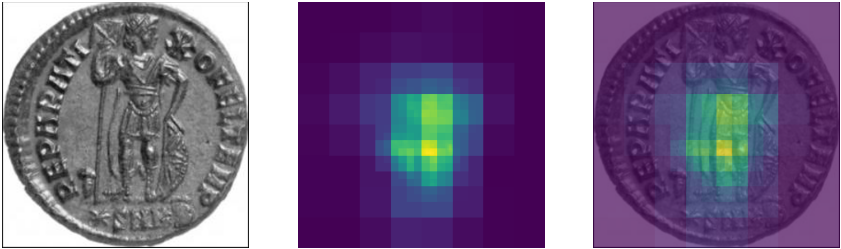}
    \end{minipage}
    \hspace{0.06\textwidth}
    \begin{minipage}{0.47\textwidth}
    \includegraphics[width=\textwidth]{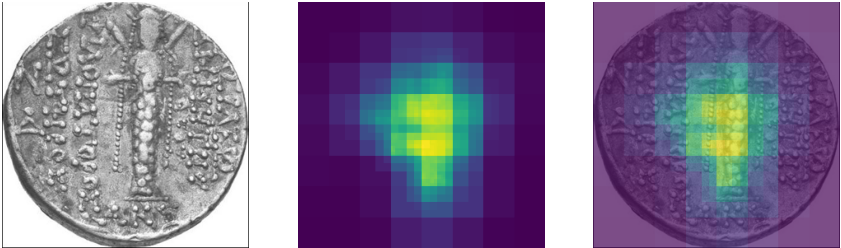}
    \includegraphics[width=\textwidth]{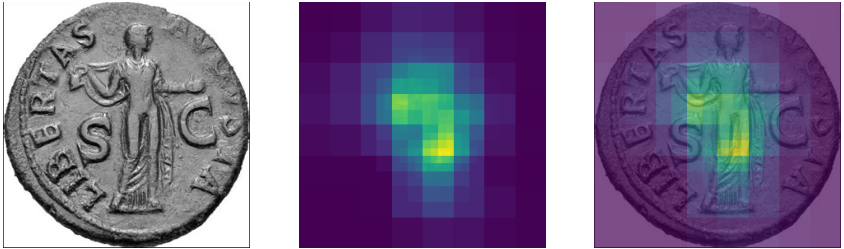}
    \end{minipage}
    \caption[CNN standing saliency maps]{The CNN model for `standing' had salient regions around the middle, generally where legs met a body.}
    \label{fig:cnn-standing}
\end{figure}

As for most other semantic elements considered, the salient regions for the ViT model were more varied than for the CNN model, which tended to focus in one area (Figure~\ref{fig:cnn-standing}).

\subsubsection{Seated}

\begin{figure}[h]
    \begin{minipage}{0.47\textwidth}
    \includegraphics[width=\textwidth]{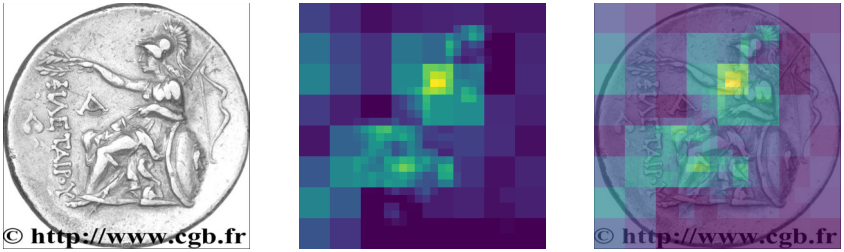}
    \includegraphics[width=\textwidth]{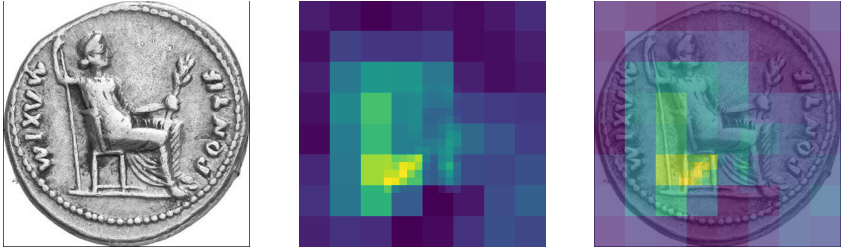}
    \end{minipage}
    \hspace{0.06\textwidth}
    \begin{minipage}{0.47\textwidth}
    \includegraphics[width=\textwidth]{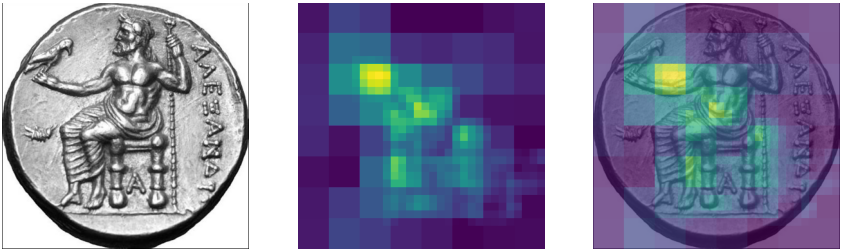}
    \includegraphics[width=\textwidth]{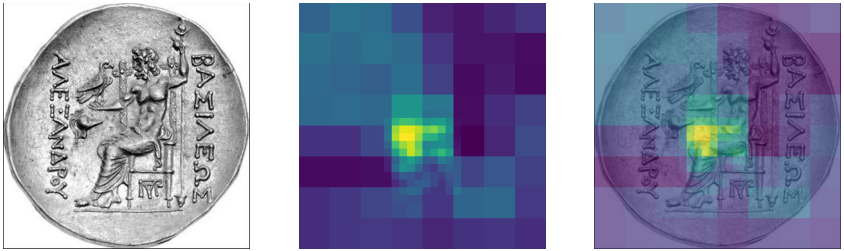}
    \end{minipage}
    \caption[ViT seated saliency maps]{It is hard to discern any obvious pattern to the salient regions of `seated' for the ViT model.}
    \label{fig:vit-seated}
\end{figure}

\begin{figure}[h]
    \begin{minipage}{0.47\textwidth}
    \includegraphics[width=\textwidth]{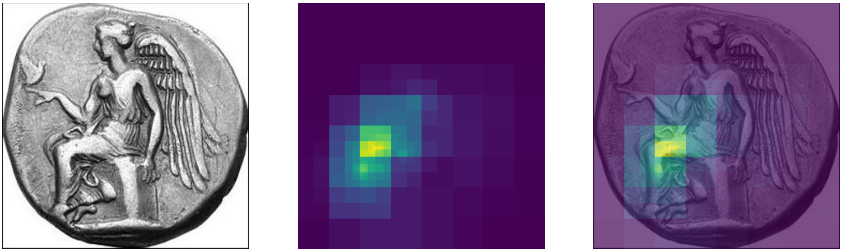}
    \includegraphics[width=\textwidth]{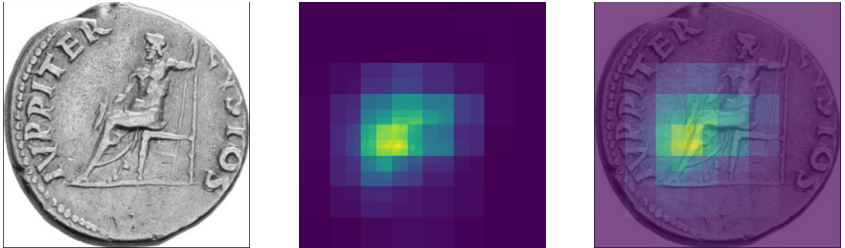}
    \end{minipage}
    \hspace{0.06\textwidth}
    \begin{minipage}{0.47\textwidth}
    \includegraphics[width=\textwidth]{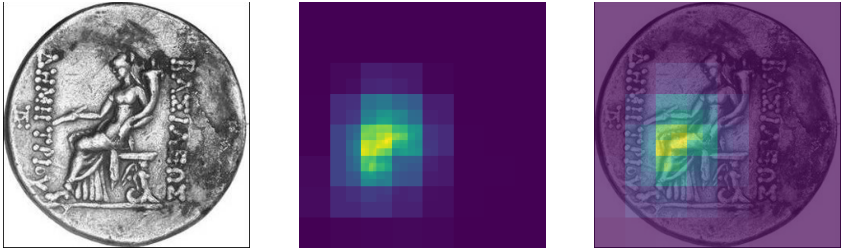}
    \includegraphics[width=\textwidth]{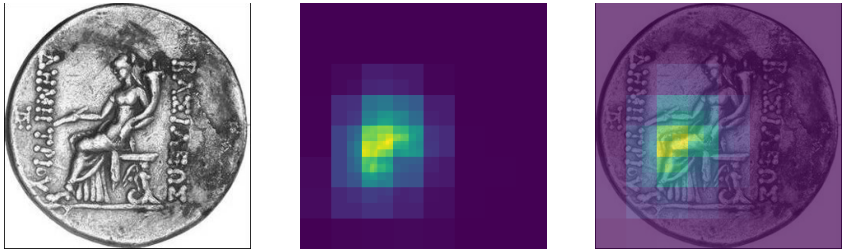}
    \end{minipage}
    \caption[CNN seated saliency maps]{The salient regions for the `seated' CNN model were consistently off to the left at around the position of seated figures' knees.}
    \label{fig:cnn-seated}
\end{figure}

As for `standing', the labels for `seated' were very noisy. A large percentage of the `false positives' for the ViT and CNN models featured seated figures that were not described as `seated' in the associated descriptions. The ViT model slightly outperformed the CNN model in terms of the test statistics. As has been the case for the other semantic elements, the ViT model's salient regions were diverse (Figure~\ref{fig:vit-seated}) and wider than the CNN model's, which were more intelligible in that they appeared to focus around the knees of figures seated facing to the left of the coin (Figure~\ref{fig:cnn-seated}).

\subsubsection{Hercules}

\begin{figure}[h]
    \begin{minipage}{0.47\textwidth}
    \includegraphics[width=\textwidth]{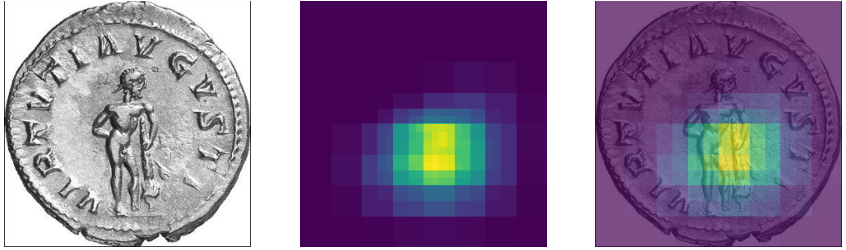}
    \includegraphics[width=\textwidth]{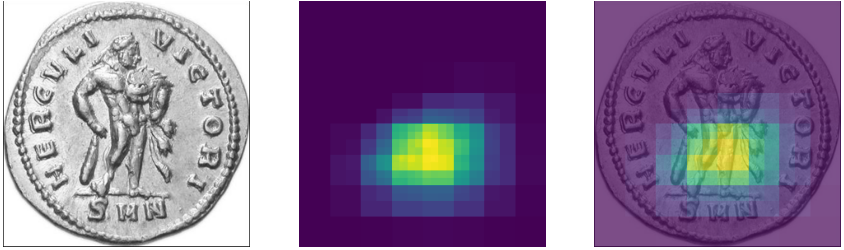}
    \end{minipage}
    \hspace{0.06\textwidth}
    \begin{minipage}{0.47\textwidth}
    \includegraphics[width=\textwidth]{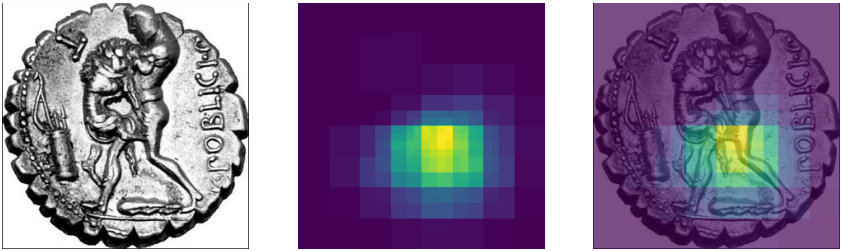}
    \includegraphics[width=\textwidth]{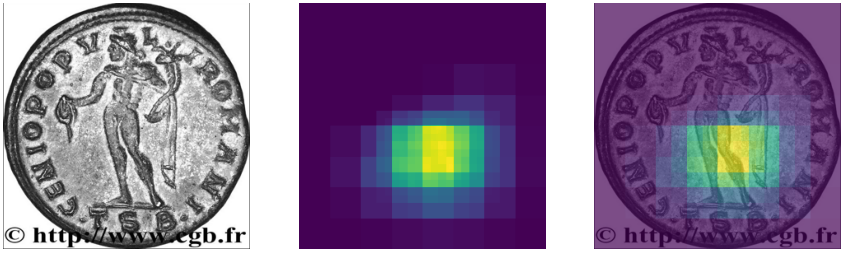}
    \end{minipage}
    \caption[CNN Hercules saliency maps]{The `Hercules' CNN model tended to focus on spot around figures' legs, suggesting that his legs must have a somewhat distinctive appearance.}
    \label{fig:cnn-hercules}
\end{figure}

 The ViT model's accuracy was joint highest for `Hercules' (as for `patera') across the semantic elements studied. The ViT model vastly outperformed the CNN model for `Hercules' (Figure~\ref{fig:cnn-hercules}): the difference in accuracy was the largest across all concepts considered. One possible factor in this is that the raw data set contained only a small number of positive samples for `Hercules' and so the balanced CNN training set, even with oversampling of the positive class, was still relatively small. The ViT model was trained on an unbalanced data set, containing far more negative samples, which may have contributed to its far superior performance overall.

\begin{figure}[h]
    \begin{minipage}{0.47\textwidth}
    \includegraphics[width=\textwidth]{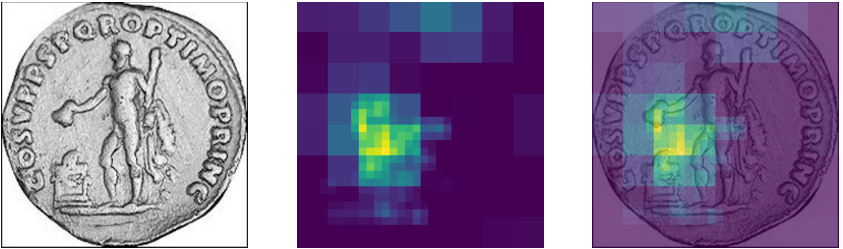}
    \includegraphics[width=\textwidth]{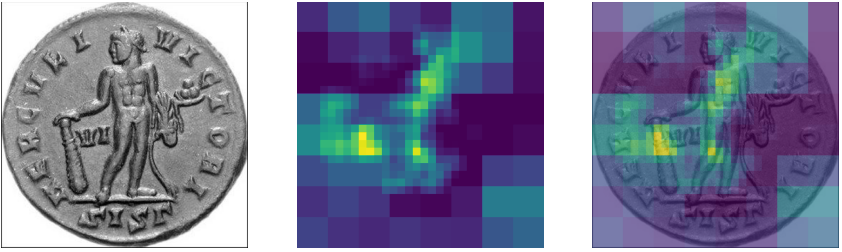}
    \end{minipage}
    \hspace{0.06\textwidth}
    \begin{minipage}{0.47\textwidth}
    \includegraphics[width=\textwidth]{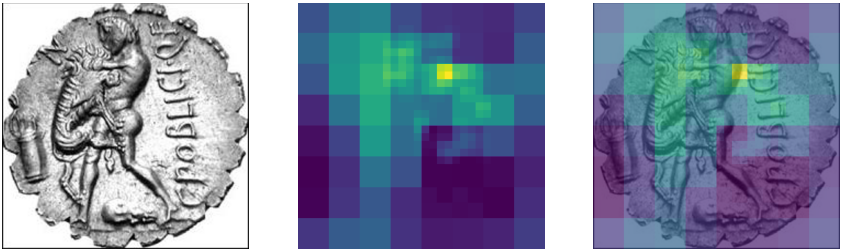}
    \includegraphics[width=\textwidth]{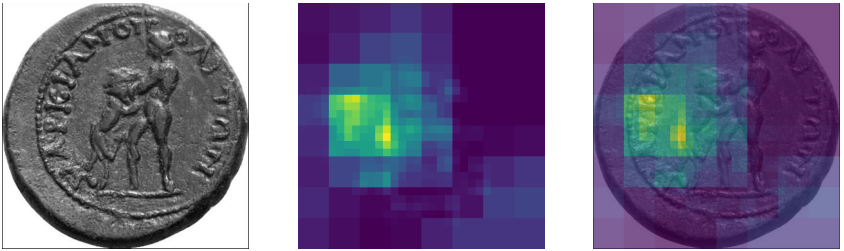}
    \end{minipage}
    \caption[ViT Hercules saliency maps]{The ViT model's salient regions for `Hercules' varied a lot, spanning letters in the legend, to objects that Hercules carried.}
    \label{fig:vit-hercules}
\end{figure}

For the ViT model (Figure~\ref{fig:vit-hercules}), a few false positives were Greek coins featuring `Herakles'---the Greek name for Hercules. When choosing to train on `Hercules', a deliberate decision was taken to ignore `Herakles' in the coin descriptions, as the Greek depiction of Herakles was often different to the Roman depiction of Hercules. It is interesting then that around 8\% of the false positives were for `Herakles'. Most of the `false negatives' for the ViT model appeared not to contain Hercules, who was described as being on the obverse side instead. Had the labelling of data been more accurate, the model could have achieved higher nominal accuracy still---that is, the aforementioned false positives are not trues error at all but rather a reflection of imperfect ground truth and an indication of the exceedingly good generalization of the model which is clearly capable of generalizing across rather substantial abstractions.

\subsubsection{Findings}

One of the recurring themes across all of the semantic elements studied is that the labels often appeared to be quite noisy, which would limit the performance of models trained using them. From the saliency maps it was apparent that the CNN models tended to focus on a small region that was relatively invariant between image samples. Although the point that a given CNN model focused on did not always seem to be the most intuitive place to look, one could at least try to look for patterns within that region to understand what cues the CNN model might be using. In contrast, the salient regions for the ViT model were more diffuse for a given image and highly variable between images depicting the same semantic element. This meant that the ViT model was quite hard to understand in any intuitive way, as there were few obvious patterns observed among the regions of saliency, even for similar images.

\subsection{Comparison to Prior Work}

The most relevant piece of prior research in this area is that of Cooper and Arandjelovi\'c \cite{cooper_learning_2020}, which was the first to focus on automatic analysis of semantic elements on coins rather than simple matching of images to coin types. This focus on identification of semantic elements has been continued in the current work, which has applied a more modern deep learning architecture (Vision Transformer) to the task and evaluated its performance in comparison to a convolutional neural network architecture, proposed for this purpose by Cooper and Arandjelovi\'c.

There were some notable differences in the results presented in this paper. The accuracy scores of the CNN models trained by Cooper and Arandjelovi\'c were significantly higher than those of the CNN models trained for this work and the time taken to train those models was significantly longer. The two differences are most likely related and probably stem from the use of a different optimization algorithm in this work. Cooper and Arandjelovi\'c used Adam as the optimizer while training their models, whereas stochastic gradient descent (SGD) was used to train the CNN models in this work. We made some attempts to train CNN models using Adam, but consistently this resulted in the so-called `dying ReLU' problem, whereby the output of large parts of the network converged to all-zeroes, leading to a model that produced a constant output independent of input. On account of this, and the fact that the ViT models had been trained using SGD, following the method of Dosovitskiy et al.\ \cite{dosovitskiy_image_2020}, the Adam optimizer that was not yielding results was swapped out for an SGD optimizer, which resulted in successful training of the CNN models.

The accuracy scores of the ViT model trained in this work are similar (within 4\% in either direction) to the accuracy scores of the CNN trained by Cooper and Arandjelovi\'c \cite{cooper_learning_2020}. This suggests that the performance of a ViT model on a data set such as the current one may not be any better than that of the CNN model previously proposed. Without further investigation, it is hard to say definitively one way or the other whether a ViT model or a CNN model is better. Further hyperparameter tuning on either architecture might result in a more accurate model. Furthermore, since Dosovitskiy et al.\ put forward the Vision Transformer \cite{dosovitskiy_image_2020}, many variants have been developed, which have their own merits. It may be that one of these other architectures would offer a significant improvement over the current ViT model or the previous CNN model.

\section{Conclusions and Future Work}\label{s:conc}

\subsection{Summary}

In this paper, we have applied the Vision Transformer (ViT) architecture to the problem of understanding the semantic elements on ancient coins for the first time. After preprocessing a data set consisting of 100,000 images and descriptions of ancient coins made available from an auction aggregator website, we trained ViT models to recognise the five semantic elements studied by Cooper and Arandjelovi\'c \cite{cooper_learning_2020}, as well as three additional concepts that were identified from a shortlist of common concepts automatically extracted from the coin descriptions. Alongside the ViT models, we trained CNN models for the same task, following the architecture proposed by Cooper and Arandjelovi\'c \cite{cooper_learning_2020}. Our evaluation found that the ViT models trained had higher accuracy than the CNN models trained, showing that their use in ancient numismatics is highly promising and that it warrants further research efforts.

\subsection{Limitations}

One of the key variables that can affect the accuracy of a deep learning model is the data set on which is is trained. In this work, the data set used, which is one of the largest used in ancient numismatics research, has some shortcomings. Namely, the image descriptions, which are used for labelling samples, relate to both sides of any coin shown in the image. Since the focus of this work was only on the semantic elements shown on the reverse side of coins, the preprocessed data set consisting of labelled images was very noisy, containing many images that were mislabelled as showing a semantic element that was in fact not visible, due to its presence on the obverse of the coin in the image. For some of the concepts that the models were trained on (e.g. `standing'), this problem was so apparent that it was likely a limiting factor in the performance of the models. Given that the ultimate purpose of this research is to advance the automated analysis of ancient coins, it must be recognised that performance of computer vision-based models for this task is going to be limited, if the data sets used for training purposes introduce a lot of noise into the process.

A further issue with the preprocessing of the data set was that the labelling process itself required as inputs a set of search words for each semantic element of interest. This process was crude and imperfect, as it would be easy to omit a word that could be indicative of the semantic element of interest (e.g. `quadriga' implies the presence of a horse, but was not included in the original list of words related to `horse'). Furthermore, the descriptions of coins covered multiple languages and so translations of each word and any derived forms had to be included. To that end, an online translation tool was used, but this could not provide a full list of all derived words that might be contained in the descriptions.

The present research built upon earlier work of Cooper and Arandjelovi\'c \cite{cooper_learning_2020}, which applied CNN models to the task of semantic analysis of ancient coins. In the present work, the results obtained from the CNN models that were trained did not match the performance of those trained by Cooper and Arandjelovi\'c, which makes it harder to draw general conclusions about the relative merits of ViT models over CNN models in this context. One confounding issue was that a different optimizer was used in the current work to the one used by Cooper and Arandjelovi\'c, due to reasons previously discussed, which may have contributed to this difference in performance.

Finally, it should be noted that the current research only explored the use of the original ViT architecture (and a particular model size and configuration thereof). Since the Vision Transformer was first proposed by Dosovitskiy et al.\ \cite{dosovitskiy_image_2020}, many related models have been put forward \cite{khan_transformers_2021}. An evaluation of architectures other than ViT and the CNN architecture previously used was outside the scope of this work.

\subsection{Future Work}

To address the confusion of descriptions of the obverse and reverse of coins in the labelling process, future work could take a few different approaches. Label smoothing could be explored as a means of coping with the noisy labels \cite{lukasik_does_2020} present in the current data set. Alternatively, the noise could be reduced by preprocessing the descriptions to filter out parts that do not relate to the coin side of interest, which could involve searching for terms like `rev' or `reverse' and their translations to identify the description boundaries. Another option would be simply to include both sides of each coin in the labelled data set, so that each semantic element in the description would at least be visible in the image, avoiding the issue of noisy labels, but at a cost of larger image sizes. Finally, future research could explore the application of multi-modal models to the current task, including multi-modal Transformers, which can learn from both images and text \cite{lu_vilbert_2019a}. These could be used in text prediction tasks to try to predict the likelihood of a description given the image, with the ultimate aim of generating a coin's description given its image.

\end{document}